\documentclass[10pt, a4paper,conference]{IEEEtran}

\usepackage{ifpdf}
\usepackage{cite}
\ifCLASSINFOpdf
   \usepackage[pdftex]{graphicx}
\else
   \usepackage[dvips]{graphicx}
\fi
\usepackage{amsmath}
\usepackage{algorithmic}
\usepackage{array}
\ifCLASSOPTIONcompsoc
 \usepackage[caption=false,font=normalsize,labelfont=sf,textfont=sf]{subfig}
\else
 \usepackage[caption=false,font=footnotesize]{subfig}
\fi
\begin{document}
%
% paper title
% Titles are generally capitalized except for words such as a, an, and, as,
% at, but, by, for, in, nor, of, on, or, the, to and up, which are usually
% not capitalized unless they are the first or last word of the title.
% Linebreaks \\ can be used within to get better formatting as desired.
% Do not put math or special symbols in the title.
\title{Sparse Network Inversion for Key Instance Detection in Multiple Instance Learning}

% author names and affiliations
% use a multiple column layout for up to three different
% affiliations
\author{\IEEEauthorblockN{Beomjo Shin, Junsu Cho, Hwanjo Yu\textsuperscript{\textsection}}
\IEEEauthorblockA{Department of CSE\\
POSTECH, Korea\\
Email: \{sbj9180, junsu7463, hwanjoyu\}@postech.ac.kr\\
}
%\and
%\IEEEauthorblockN{Junsu Cho}
%\IEEEauthorblockA{Department of CSE\\
%POSTECH, Korea\\
%Email: junsu7463@postech.ac.kr}
%\and
%\IEEEauthorblockN{Hwanjo Yu}
%\IEEEauthorblockA{Department of CSE\\
%POSTECH, Korea\\
%Email: hwanjoyu@postech.ac.kr}
\and
\IEEEauthorblockN{Seungjin Choi}
\IEEEauthorblockA{Inference Lab\\
BARO AI, Korea\\
Email: seungjin@baroai.com\\
}}

% conference papers do not typically use \thanks and this command
% is locked out in conference mode. If really needed, such as for
% the acknowledgment of grants, issue a \IEEEoverridecommandlockouts
% after \documentclass

% for over three affiliations, or if they all won't fit within the width
% of the page, use this alternative format:
%
%\author{\IEEEauthorblockN{Michael Shell\IEEEauthorrefmark{1},
%Homer Simpson\IEEEauthorrefmark{2},
%James Kirk\IEEEauthorrefmark{3},
%Montgomery Scott\IEEEauthorrefmark{3} and
%Eldon Tyrell\IEEEauthorrefmark{4}}
%\IEEEauthorblockA{\IEEEauthorrefmark{1}School of Electrical and Computer Engineering\\
%Georgia Institute of Technology,
%Atlanta, Georgia 30332--0250\\ Email: see http://www.michaelshell.org/contact.html}
%\IEEEauthorblockA{\IEEEauthorrefmark{2}Twentieth Century Fox, Springfield, USA\\
%Email: homer@thesimpsons.com}
%\IEEEauthorblockA{\IEEEauthorrefmark{3}Starfleet Academy, San Francisco, California 96678-2391\\
%Telephone: (800) 555--1212, Fax: (888) 555--1212}
%\IEEEauthorblockA{\IEEEauthorrefmark{4}Tyrell Inc., 123 Replicant Street, Los Angeles, California 90210--4321}}

% use for special paper notices
%\IEEEspecialpapernotice{(Invited Paper)}

% make the title area
\maketitle
\begingroup\renewcommand\thefootnote{\textsection}
\footnotetext{Corresponding author}
\endgroup

% As a general rule, do not put math, special symbols or citations
% in the abstract
\begin{abstract}
Multiple Instance Learning (MIL) involves predicting a single label for a bag of instances, given positive or negative labels at bag-level, without accessing to label for each instance in the training phase. Since a positive bag contains both positive and negative instances, it is often required to detect positive instances (key instances) when a set of instances is categorized as a positive bag. The attention-based deep MIL model is a recent advance in both bag-level classification and key instance detection (KID). However, if the positive and negative instances in a positive bag are not clearly distinguishable, the attention-based deep MIL model has limited KID performance as the attention scores are skewed to few positive instances. In this paper, we present a method to improve the attention-based deep MIL model in the task of KID. The main idea is to use the neural network inversion to find which instances made contribution
to the bag-level prediction produced by the trained MIL model.
Moreover, we incorporate a sparseness constraint into the neural network inversion, leading to
the {\em sparse network inversion} which is solved by the proximal gradient method.
Numerical experiments on an MNIST-based image MIL dataset and two real-world histopathology datasets
verify the validity of our method, demonstrating the KID performance is significantly improved while the performance of bag-level prediction is maintained.

\end{abstract}

% no keywords

\IEEEpeerreviewmaketitle

\section{Introduction}
\label{sec:introduction}

Machine learning has successfully solved typical classification problems where a class label is assigned to each data instance. However, in many real-world classification problems, multiple instances are observed and only a general class label for these instances are given \cite{Ilse2018icml}. Such classification problems are called multiple instance learning (MIL) problem \cite{Dietterich97, Maron98}. MIL problems are commonly found in read-world scenarios, such as medical field \cite{queelcArticle}. For example, a medical image can consist of several image patches, each of which correspond to benign or malignant. The label of the image is determined by the presence of the malignant patch. These MIL problems are necessary because labeling every data instance requires a lot of costs (labor and time) from experts \cite{queelcArticle}.

%These MIL problems are commonly found in medical images \cite{queelcArticle}. 
MIL has traditionally assumed that the bag-level label is positive if at least one of the instances in a bag is a positive instance, otherwise negative, and there are no dependency and no ordering between instances in a bag \cite{zhouiid}. Under the assumption, MIL learns the relationship between a set of instances called bag and bag-level label. In addition to learning the mapping between bag and bag-level label, MIL also aims to find the positive instances called key instances that trigger the bag-level label without accessing instance-level label \cite{Liu2012KeyID}. These two tasks of the MIL are called bag-level classification and key instance detection (KID), respectively. 

% check
The conventional methods for bag-level classification and KID in MIL are based on instance-space paradigm \cite{amores} that trains an instance-level classifier \cite{miSVM, WANG201815, Ramn2000MultiIN}. The methods have performance limits in bag-level classification. The reason is that the instance-level classifier is not sufficiently trained as the data for MIL do not contain instance-level labels \cite{Ilse2018icml}. To solve this problem, an attention-based deep MIL model is proposed \cite{Ilse2018icml}, which is based on embedding-space paradigm \cite{amores}. The attention-based deep MIL model transforms the instances in a bag to instance embeddings, and then makes a bag embedding by aggregating the instance embeddings through attention-based pooling. Finally, the MIL model calculates a bag-level score by using a bag-level classifier that takes the bag embedding as input. This kind of MIL model has better bag-level classification performance than the methods based on instance-space paradigm because the model utilizes the bag-level data and label to train the bag-level classifier, not the instance-level classifier. Moreover, the attention-based deep MIL model can perform KID by using the attention scores of the instances in a bag when the bag-level label is positive.

Despite the superiority of the bag-level classification performance of the attention-based deep MIL model, the model has performance limits in terms of KID. The reason is that the model cannot learn every key instance in detail as the model uses only the bag-level data and bag-level label without accessing instance-level data and instance-level label. Thus, the attention-based deep MIL model focuses only on few distinguishable key instances that trigger a bag-level label, and then the indistinguishable key instances are treated as negative instances by the model. 

In the paper, we propose a novel method to improve KID performance of the attention-based deep MIL model while maintaining the bag-level classification performance. Our method consists of two modules: (1) a trained MIL module to make a bag-level prediction; (2) neural network inversion with a sparseness constraint (sparse network inversion) module. For the first module, an attention-based deep MIL model is trained by optimizing negative log-likelihood function for bag-level label. In the second module, whenever the trained MIL model classifies the input bag as positive, the sparse network inversion changes the values of instances in the bag to fit the criterion of the MIL model. In addition to optimizing the bag to fit the criterion of the trained MIL model, parts of the optimized bag that do not affect the criterion are weakened by the effect of a sparseness constraint. That is, as a result of sparse network inversion, the parts of the optimized bag that make the bag positive are strengthened, and the other parts are weakened. Therefore, if the optimized bag is put to the trained MIL model then, the attention scores corresponding to the key instances become larger than before applying sparse network inversion to the bag. As a result, our method relieves the problem of the attention-based deep MIL model by removing the constraint that input data cannot be changed, so the KID performance of the attention-based deep MIL model is significantly improved. In the experiments, we show that our method significantly improves the KID performance with the superiority of bag-level classification performance in the attention-based deep MIL model for an MNIST-based image MIL dataset and two real-world histopathology datasets.

\section{Related Work}
\label{sec:related}

\subsubsection{Multiple Instance Learning and Key Instance Detection}

MIL is a type of typical weakly supervised learning. Since MIL is relatively free from the cost of labeling data compared to standard supervised learning, the methods for classification of multiple instances have been actively studied after it was first proposed for Drug Activity Prediction in 1997 \cite{Dietterich97, miSVM, Ilse2018icml}. KID is a task of MIL, which detects key instances in a positive bag \cite{Liu2012KeyID}. KID is more challenging than bag-level classification \cite{Liu2012KeyID}, as the MIL models are trained only with bag-level labels without accessing instance-level labels. Despite its difficulty, KID is considered as an important task and many methods for KID have been developed \cite{Liu2012KeyID, Wang2016AMI, Ilse2018icml} as the bag-level decision of the MIL model can be interpreted with the key instances that make the bag-level label positive. The interpretation of bag-level decision is important in practical applications. For example, in the computational pathology, providing both a final diagnosis and the basis of the diagnosis is more useful than providing only diagnosis about the image \cite{Ilse2018icml}.

The conventional methods for bag-level classification and KID include the MIL methods based on instance-space paradigm and Voting Framework (VF) solution \cite{miSVM, Wang2016AMI, Liu2012KeyID}. The MIL methods based on instance-space paradigm train an instance-level classifier to calculate the score of each instance in a bag for instance-level classification. Then, the responses of the instance-level classifier are aggregated for bag-level classification. These methods have a worse bag-level classification performance than the models with embedding-space paradigm, as the instance-level classifier is trained only with the bag-level data and label \cite{Ilse2018icml}. The VF solution is a method of using neighborhood relations among instances. Therefore, this method can be effective only when a proper neighborhood relationship among instances are obtained \cite{Liu2012KeyID}, whereas the method is difficult to apply in many real situations. In order to solve these problem, the methods that apply interpretable attention-based pooling or dynamic pooling to embedding-space paradigm have recently been proposed \cite{Ilse2018icml, Yan2018DeepML}. These methods use the bag embedding to classify a bag. Since only bag-level information is used for training the MIL model without accessing the instance-level information, these methods have better bag-level classification performance than the MIL models based on instance-space paradigm. However, the MIL model using dynamic pooling is not suitable for situations where the instances in a bag are independent as the model utilizes the contextual information among the instances in the bag. In the case of the MIL model using attention-based pooling, the model has performance limits in terms of KID as the MIL model cannot learn every key instance in detail.

\begin{figure*}
\centering
\includegraphics[width=1.0\linewidth]{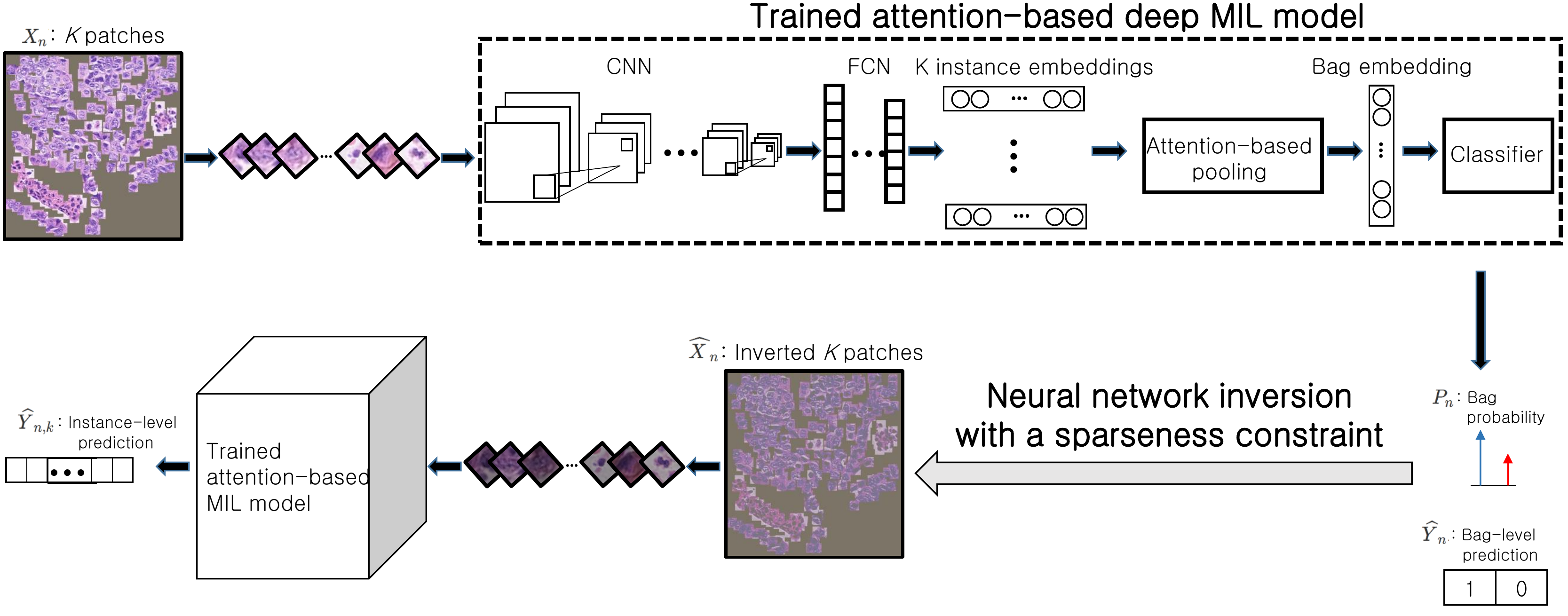}
\caption{The flow diagram of the proposed method.}
\label{fig:main}
\end{figure*}

\subsubsection{Neural Network Inversion}

Neural network inversion is a method that optimizes the input to find an input pattern that fits the criterion of the trained neural network model \cite{Kindermann1990InversionON}. Neural network inversion consists of two steps: the first step is to train the neural network to optimize its criterion, and the second step is to optimize the input variables of the trained neural network \cite{Hoskins:1992:IIN:2325821.2327686, Kindermann1990InversionON}. Neural network inversion is used to find optimal input(s) in various industrial fields, such as a plant or a device for a heating, ventilation, and air conditioning (HVAC) system of a building \cite{Hoskins:1992:IIN:2325821.2327686, JANG2020229}. 

For bag-level classification, we use the attention-based deep MIL model. For KID, to solve the problem of the attention-based deep MIL model, we apply a neural network inversion with a sparseness constraint that updates the instances in a positive bag. That is, our method removes the constraint on data, so that the optimized instances can be changed to fit the criterion of the model, and the model can easily find out the key instances by using the optimized instances. To the best of our knowledge, our method is the first attempt that applies neural network inversion to the MIL model.

\section{The Proposed Method}
\label{sec:main}

Suppose that we are given a set of images $\{ X_n \,|\, n=1,\ldots,N\}$, each of which contains $K$ image patches.
That is, an image $X_n$ is treated as a bag of $K$ instances. 
The number of patches, $K$, can vary depending on images.
Each image in the training set is associated with a label $Y_n \in \{0,1\}$ ($Y_n=1$ indicates that
$X_n$ is a positive bag) at the bag level,
without the knowledge of an instance-level label $Y_{n,k}$.

The flow diagram of our method is shown in Fig. \ref{fig:main},
where an input image $X_n$ is applied to the method which constitutes two modules:
(1) an already-trained MIL module to yield a bag-level prediction;
(2) a sparse network inversion module.
We use the attention-based deep MIL model \cite{Ilse2018icml} that is already trained using the training set.
Our contribution lies at the {\em sparse network inversion} to refine 
the input image, finding instances that make contributions to the bag-level prediction $\widehat{Y}_n$ 
made by the trained MIL module.
The MIL module is first trained using the training set.
Then, whenever $\widehat{Y}_n=1$, the sparse network inversion is applied to refine the input image,
leading to $\widehat{X}_n$.
The refined input image $\widehat{X}_n$ is applied to the MIL module again to recompute the
attention scores that are used to finally detect key instances.

%-------------------------------------------------------------------------

\subsection{Problem Formulation} MIL deals with a set of bags $\{ X_n \,|\, n=1,\ldots,N\}$, where $N$ is the number of bags. A bag $X_n$ is associated with multiple instances $\{ X_{n,k} \in {\rm I\!R}^D \,|\, k=1,\ldots,K\}$, where $K$ is the number of instances ($K$ can vary for different bags), and there are no dependency and no ordering between $K$ instances in a bag. Each instance in the bag $X_n$ is associated with an instance-level label $\{ Y_{n,k} \in \{0,1\} \,|\, k=1,\ldots,K\}$, and the bag $X_n$ is associated with the bag-level label $Y_n = OR(Y_{n,1}, \ldots, Y_{n,K})$. That is, the bag-level label is positive if at least one of the instances in the bag is a positive instance, otherwise negative. Under the MIL assumption, an attention-based deep MIL model computes a bag probability $P_n$ and attention scores $a_n$ for the $n^\text{th}$ bag $X_n$, where $a_{n}$ is a set of attention scores of $K$ instances in the $n^\text{th}$ bag, i.e., $a_{n}=\{a_{n,1}, \ldots ,a_{n,K}\}$. For the $n^\text{th}$ bag $X_n$, the bag-level prediction $\widehat{Y}_{n}$ is computed by applying a threshold to bag probability $P_n$. In the case of KID, whenever $\widehat{Y}_{n}=1$, for the $k^\text{th}$ instance $X_{n,k}$ in the $n^\text{th}$ bag, an instance-level prediction $\widehat{Y}_{n,k}$ is computed by applying a threshold to the normalized attention score $\widehat{a}_{n,k}$ after Min-Max normalization is applied to the attention score $a_{n,k}$.

%-------------------------------------------------------------------------

\subsection{Attention-based deep MIL model}

We build the attention-based deep MIL model that uses attention-based pooling on embedding-space paradigm as shown in Fig. 1, and we train the model by optimizing negative log-likelihood function. There are two reasons we use this structure for the MIL model. First, embedding-space paradigm has better bag-level classification performance than instance-space paradigm. The second reason is that attention-based pooling allows embedding-space paradigm to detect key instances because the magnitude of the attention score corresponding to an instance in a positive bag tells how likely it is to be a key instance.

As shown in the Fig. 1, the attention-based deep MIL model transforms the instances in the bag $X_n$ to instance embeddings, and then the model makes a bag embedding by aggregating the instance embeddings through attention-based pooling. Finally, the model calculates a bag probability $P_n$ by putting the bag embedding to a bag-level classifier. For the MIL model, if we assume that $G_n = \{g_{n,1}, \ldots, g_{n,K}\}$ is a set of the instance embeddings of the bag $X_n$, where $g_{n,k}$ is the instance embedding of the $k^\text{th}$ instance in the $n^\text{th}$ bag and $g_{n,k} \in $ {\rm I\!R}$^{1 \times N}$, an attention score for the $k^\text{th}$ instance in the $n^\text{th}$ bag can be written as follows \cite{Ilse2018icml}: 

\begin{equation} 
a_{n,k} = \frac{\exp\left\{\text{w}^\top \text{tanh}(\text{V}g_{n,k}^\top)\right\}}{\sum_{j=1}^K\exp\left\{\text{w}^\top \text{tanh}(\text{V}g_{n,j}^\top)\right\}}
\label{equation2}\end{equation}
where $a_{n, k}$ is an attention score for the $k^\text{th}$ instance in the $n^\text{th}$ bag, $\text{w} \in ${\rm I\!R}$^{M \times 1}$ and $\text{V} \in ${\rm I\!R}$^{M \times N}$ are fully-connected layer's parameters. The bag embedding for the $n^\text{th}$ bag based on attention-based pooling can be written as follows \cite{Ilse2018icml}: 

\begin{equation} 
z_n=\sum_{j=1}^Ka_{n,j}g_{n,j}
\label{equation3}\end{equation}
where $z_n$ is the bag embedding for the $n^\text{th}$ bag. The bag embedding $z_n$ is put to a bag-level classifier $H\left(\cdot\right)$, and then the bag probability is calculated by using the result of bag-level classifier $H\left(\cdot\right)$. This process can be written as follows: 

\begin{equation} 
P_n = \text{sigmoid}\left(H \left(z_n\right)\right)
\label{equation4}
\end{equation}
where $P_n$ is the bag probability for the $n^\text{th}$ bag. For the bag-level classifier $H\left(\cdot\right)$, we use a single fully-connected layer, which can learn the interactions within the bag embedding. To make the bag-level prediction $\widehat{Y}_n$, we apply a threshold to $P_n$. In the case of KID, whenever $\widehat{Y}_n = 1$, we apply Min-Max normalization to the attention score $a_{n,k}$ of $k^\text{th}$ instance in the bag $n^\text{th}$ bag. This process can be written as follows:

\begin{equation} 
\widehat{a}_{n,k} = \frac{a_{n,k} - \min(a_n)}{\max(a_n)-\min(a_n)}
\label{equationminmax}\end{equation}
where $\widehat{a}_{n,k}$ is the normalized attention score of $a_{n,k}$ and $a_n$ is a set of attention scores of $K$ instances in the $n^\text{th}$ bag, i.e., $a_{n}=\{a_{n,1}, \ldots ,a_{n,K}\}$. Then, we make instance-level prediction $\widehat{Y}_{n,k}$ by applying a threshold to $\widehat{a}_{n,k}$. However, since the attention-based deep MIL model has limits in KID performance, we perform KID by recomputing the instance-level prediction $\widehat{Y}_{n,k}$ of the optimized bag $\widehat{X}_n$ after optimizing the bag $X_n$ to fit the criterion of the model by applying sparse network inversion.

%-------------------------------------------------------------------------
\subsection{Neural Network Inversion}

Neural network inversion is a method that optimizes the input to fit the criterion of the neural network model. To apply neural network inversion to an attention-based deep MIL model, we initialize the input as the values of the input image. Since we use negative log-likelihood as objective function when training the attention-based deep MIL model, we optimize the bag $X_n$ by fitting the same criterion of the attention-based deep MIL model. That is, whenever the bag-level prediction $\widehat{Y}_{n}$ for the bag $X_n$ is positive, we apply neural network inversion to the bag $X_n$. Then, the parts of the optimized bag $\widehat{X_n}$ that make the bag positive are strengthened, and the other parts are weakened. In other words, if the optimized bag $\widehat{X_n}$ is put to the trained MIL model again, the attention scores corresponding to the key instances become larger. Formally, the objective function of neural network inversion can be written as follows:

\begin{equation}
l(X_n) = -\widehat{Y}_n \log P_n - \left(1-\widehat{Y}_n \right) \log(1-P_n)\ 
\label{equation7}
\end{equation}
where $\widehat{Y}_{n}$ is bag-level prediction for the bag $X_n$. The reason we use $\widehat{Y}_{n}$, not $Y_n$, is that we cannot access the true bag-level label $Y_n$ during test. To update the bag $X_n$ by optimizing the objective function of neural network inversion, we use Stochastic Gradient Descent (SGD) optimization algorithm \cite{ruder2016overview}. Each time the instances of the bag $X_n$ is updated, we set the range of the values of instances from $0$ to $1$ as each instance is image data with a value between $0$ and $1$. 

\subsection{Sparse Network Inversion}

Sparse network inversion is a kind of neural network inversion that uses a sparseness constraint. That is, sparse network inversion updates the bag $X_n$ by optimizing the objective function of neural network inversion with a sparseness constraint. We incorporate a sparseness constraint into neural network inversion to give regularization effect when we update the bag $X_n$ and to further weaken the parts of the bag $X_n$ that do not affect the criterion of the trained MIL model. Thus, if we put the optimized bag $\widehat{X}_n$ to the attention-based deep MIL model, we can detect more key instances than when using only attention-based deep MIL model or applying neural network inversion. The objective function of sparse network inversion can be written as follows:  

\begin{equation}
L(X_n) = l(X_n) + \lambda \left\lVert  X_n\right\rVert _\text{1}
\label{equation8}
\end{equation} 
where $l(X_n)$ is the objective function of neural network inversion and $\lambda$ is a sparseness coefficient. Since the objective function of sparse network inversion involves a sparseness constraint for bag $X_n$, in order to optimize the bag $X_n$, we use \textit{proximal gradient method}, which is an extension of typical gradient algorithm \cite{ista}. The method is used due to its simplicity and adequateness for solving data with high dimension \cite{ista, prox}. In the proximal gradient method, soft-thresholding function is used as proximal operator for a sparseness constraint. The soft-thresholding function can be written as follows \cite{ista}:

\begin{equation}
s_\lambda(x)=
\begin{cases}
x-\lambda & \mbox{if }x > \lambda\\
0 & \mbox{if }\left\vert x \right\vert \le \lambda\\
x+\lambda & \mbox{if }x < -\lambda
\end{cases}
\label{equation9}
\end{equation}
where $\lambda$ is the sparseness coefficient. We can write the soft-thresholding function compactly as follows:
\begin{equation}
s_\lambda(x)= \max (\left\vert x \right\vert - \lambda,0) \mbox{ sign($x$)}
\label{equation10}
\end{equation}
where sign($x$) means the sign of $x$. Formally, when we use proximal gradient method for the bag $X_n$, update equation of sparse network inversion can be written as follows:
\begin{equation}
X_n^{t+1} = s_\lambda(X_n^t - \eta \nabla l(X_n^t))
\label{equation11}
\end{equation}
where $s_\lambda(\cdot)$ is the soft-thresholding function, $\eta$ is the learning rate, and $l(\cdot)$ is the objective function of neural network inversion. Since instances of the bag $X_n$ is image data with a value between $0$ and $1$, each time each instance of the bag $X_n$ is updated, we set the range of the instances in the bag from $0$ to $1$. 

\section{Experiments}

In this section, we introduce an experimental setting, datasets, implementation details, and performance results and analyses of our proposed method. Through the section, we aim to show that our method improves the performance in terms of KID with the superiority of the bag-level classification performance of the attention-based deep MIL model \cite{Ilse2018icml}. We evaluated our method on various MIL datasets including an MNIST-based image MIL dataset, and two real-world histopathology datasets (COLON CANCER \cite{sir}, BREAST CANCER \cite{breast}). Since our method is the same as the attention-based deep MIL model for the bag-level classification, to show the superiority of our proposed method in bag-level classification performance, we compare the attention-based deep MIL model with the MIL models based on instance-space paradigm. Since KID detects the positive instances that triggered a bag-level label in a positive bag, KID performance becomes meaningful only if the MIL model can classify the bags properly. Thus, we compare our proposed method with the attention-based deep MIL model and the MIL model that uses max pooling on instance-space paradigm, which showed best bag-level classification performance among the MIL models based on instance-space paradigm in our experiments.

\begin{table}[!t]
\caption{Percentage of bags on various MIL datasets. Positive (Negative) presents the percentage of positive (negative) bags in each dataset.}
\label{tab:bags}
\centering
\begin{tabular}{|c||c|c|}
\hline
\bfseries Dataset  & \bfseries Positive & \bfseries Negative \\
\hline
MNIST-based image MIL & 50\% & 50\% \\
\hline
COLON CANCER & 51\% & 49\% \\
\hline
BREAST CANCER & 44.8\% & 55.2\% \\
\hline
\end{tabular}
\end{table}

\begin{table}[!t]
\caption{Percentage of instances in positive bags on various MIL datasets. Positive (Negative) presents the percentage of positive (negative) instances in each dataset.}
\label{tab:instances}
\centering
\begin{tabular}{|c||c|c|}
\hline
\bfseries Dataset  & \bfseries Positive & \bfseries Negative \\
\hline
MNIST-based image MIL & 10.2\% & 89.8\% \\
\hline
COLON CANCER & 53.5\% & 46.5\% \\
\hline
BREAST CANCER & 8.5\% & 91.5\% \\
\hline
\end{tabular}
\end{table}

\subsection{Experimental Setting}

For all experiments, we use the same model architecture and optimizer that showed the high classification performance for each dataset as used in the precedent work \cite{Ilse2018icml}. To construct the attention-based deep MIL model, we apply attention-based pooling before the last layer of the deep neural network model. We call the attention-based deep MIL model Att, the method that applies neural network inversion to the attention-based deep MIL model Att+inv, and the method that applies sparse network inversion to the attention-based deep MIL model Att+sparse. In the MIL models based on instance-space paradigm, the models compute instance scores and aggregate the scores by max pooling or mean pooling. To construct the model, we applied max pooling or mean pooling after the last layer of the deep neural network model. We call these MIL models Inst+max, Inst+mean. 

To do a fair evaluation, we use 10-fold cross-validation, and each experiment was performed five times independently. In the case of the MNIST-based image MIL dataset, we divided the MNIST dataset into an MNIST train dataset and an MNIST test dataset. Then, we created the MNIST-based image MIL train dataset and the MNIST-based image MIL test dataset by doing sampling from an MNIST train dataset and an MNIST test dataset independently. 

To compare the bag-level classification performance, we use accuracy as a metric because the number of positive and negative bags on various MIL datasets is balanced as shown in Table 1. In the case of KID performance, since the number of positive and negative instances in positive bags on various MIL datasets are imbalanced as shown in Table 2, we use F1 measure as used in the precedent work \cite{Liu2012KeyID, instf1}, as F1 measure is known to be insensitive to imbalance of the labels in data \cite{f1measure}. 

\subsection{Datasets} 

\subsubsection{MNIST-based image MIL dataset}

To intuitively show that our method improves the KID performance while maintaining the bag-level classification performance of the attention-based deep MIL model, we created an MNIST-based image MIL dataset using the MNIST dataset. Each image of the MNIST-based image MIL dataset is shown in Fig. 3-(a). To make each data of the MNIST-based image MIL dataset that looks like an image as shown in Fig. 3-(a), we randomly extracted 400 images from the MNIST dataset and arranged them by $20\times20$. That is, each data in the MIL dataset consists of 400 MNIST images as shown in Fig. 3-(a), where each MNIST image in a white box is an instance in a bag, and a set of 400 MNIST images is the bag. The label of each bag is assigned as a positive label if one image or more images of 400 MNIST images correspond to number 9, otherwise a negative label. The reason we chose the number 9 as the basis for making the bag label is that using the number 9 makes the problem difficult as the number 9 is easily confused with other numbers \cite{Ilse2018icml}. 

\subsubsection{Histopathology datasets}

To verify the superiority of our method, we conducted an experiment with weakly labeled real-world histopathology datasets including colon cancer dataset (COLON CANCER) \cite{sir} and breast cancer dataset (BREAST CANCER) \cite{breast}. 

COLON CANCER consists of 100 H\&E stained histology images. Each image consists of 500$\times$500 pixels. In the dataset, there are 29,756 nuclei that were marked. Among these nuclei, 22,444 nuclei have an associated class label, i.e., epithelial, inflammatory, fibroblast, and miscellaneous. To use this dataset in MIL, as shown in Fig. 4-(a), we created the bag with 27$\times$27 patches as used in the precedent work \cite{Ilse2018icml}. Each of the bags is labeled as positive if it contains one or more nuclei associated with the epithelial class, otherwise negative \cite{Ilse2018icml, sir}. The reason we choose epithelial cells as the key instances that trigger bag-level label is that colon cancer is closely related to epithelial cells \cite{epti}.

BREAST CANCER consists 58 H\&E images. Each image consists of 896$\times$768 pixels. To use this dataset in MIL, we made the bags with 32$\times$32 patches as used in precedent work \cite{Ilse2018icml}. When we make bags, if each patch  contains 75$\%$ or more white pixels, we discarded the patch. Each of the bags is labeled as positive if it contains one or more cancer cells, otherwise negative \cite{Ilse2018icml, breast}.

\subsection{Implementation details}

To experiment with these datasets, we used an existing model architectures: Lenet 5 model \cite{Lenet} for MNIST-based image MIL dataset; the proposed model \cite{sir} for two real-world histopathology datasets. When constructing the models, ReLU is used as activation function in all layers except for the two layers that is used for attention-based pooling, where tanh is used as the activation function \cite{Ilse2018icml}. To optimize the MIL models, Adam optimization algorithm is used \cite{adam}. For the MIL models on MNIST-based image MIL dataset, beta values of 0.9 and 0.999, learning rate of 0.0005, and weight decay of 0.0001 are used \cite{Ilse2018icml}. In the two histopathology datasets, beta values of 0.9 and 0.999, learning rate of 0.0001, weight decay of 0.0005 are used \cite{Ilse2018icml}. When optimizing the models, we used 100 epochs and we stopped optimizing models based on the lowest validation error. Also, in the case of two histopathology datasets, to prevent overfitting due to the small amount of data, we performed data augmentation by randomly rotating and flipping the data. In order to improve KID performance of the attention-based deep MIL model, when we optimize data by applying neural network inversion, we updated each data 1,000 times by using SGD optimization algorithm \cite{ruder2016overview} where learning rate is 0.001, momentum coefficient is 0.9. When we applied sparse network inversion, we used proximal gradient method. For MNIST-based image MIL dataset, we used 0.0001 as learning rate and we repeated update 200 times with varying the sparseness constraint coefficient from 0.0005 to 0.003. In COLON CANCER, the learning is 0.00001 and each data is updated 200 times with varying the sparseness constraint coefficient from 0.0004 to 0.0024. In the case of BREAST CANCER, 0.0001 is used as learning rate and each data is updated 100 times with varying the sparseness constraint coefficient from 0.003 to 0.006. For KID, we set the threshold to the optimal value among [0.1, 0.2, 0.3, 0.4, 0.5] for each model and dataset.

\begin{table}[!t]
\caption{Bag-level classification accuracy on various MIL datasets. All results are averages of 5 times running and ($\pm$) is a standard error of the mean.}
\label{tab:booktabs}
\centering
\resizebox{\columnwidth}{!}{
\begin{tabular}{|c||c|c|c|}
\hline
\bfseries Method  & \bfseries MNIST-based image MIL & \bfseries COLON CANCER & \bfseries BREAST CANCER \\
\hline
Inst+max       & 0.804 $\pm$ 0.232 & 0.868 $\pm$ 0.025 & 0.536 $\pm$ 0.062     \\
\hline
Inst+mean       & 0.708 $\pm$ 0.095 & 0.798 $\pm$ 0.023& 0.612 $\pm$ 0.038    \\
\hline
Attention       & 0.996 $\pm$ 0.008 & 0.909 $\pm$ 0.02 & 0.718 $\pm$ 0.054    \\
\hline
\end{tabular}
}
\end{table}

\begin{figure}
\centering
\subfloat[F1 measure for an MNIST-based image MIL dataset.]{\includegraphics[width=1.0\linewidth]{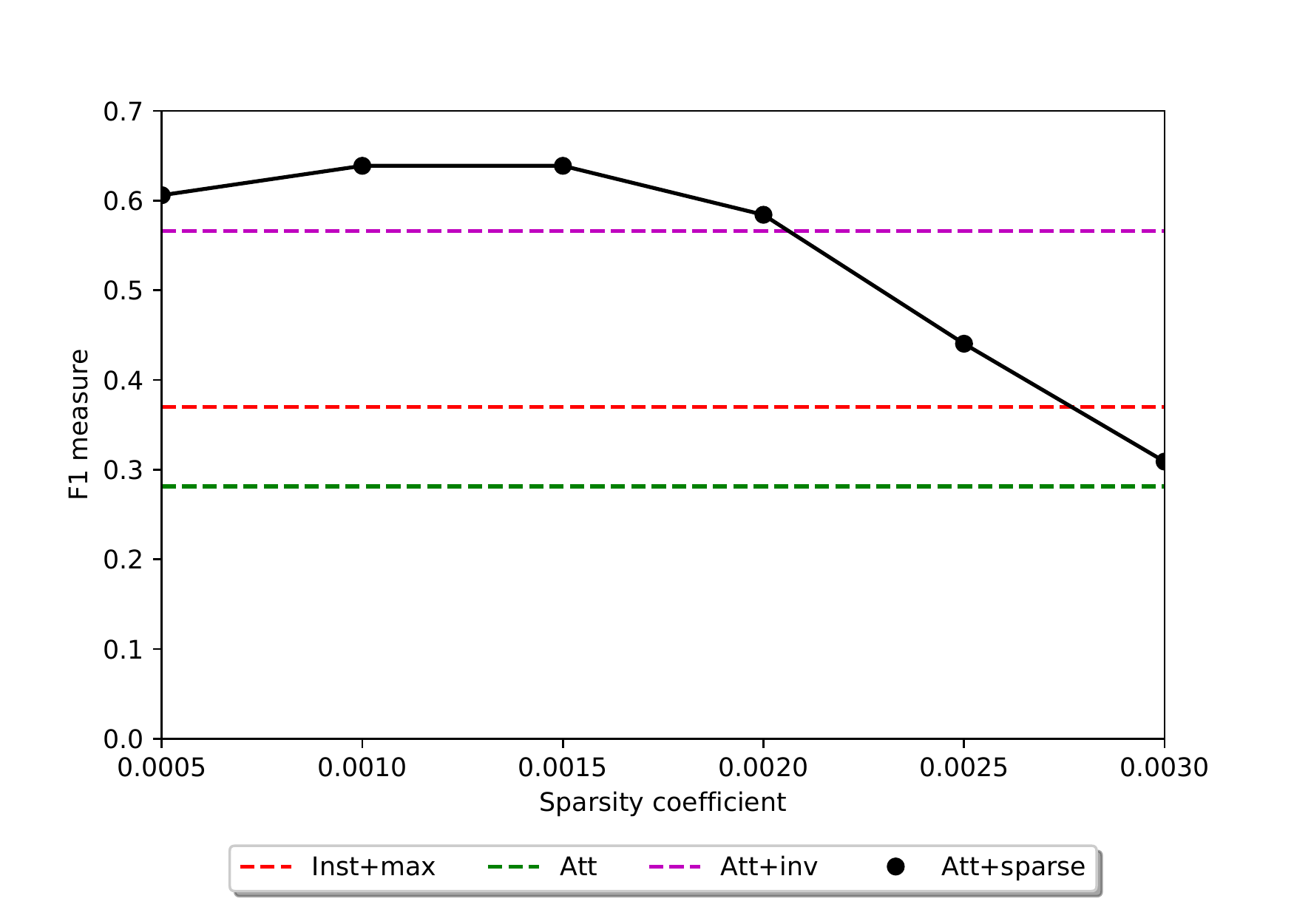}%
% \caption{Simulation results for the network.}
\label{mnist:f1}}

\hfil
\subfloat[F1 measure for COLON CANCER.]{\includegraphics[width=1.0\linewidth]{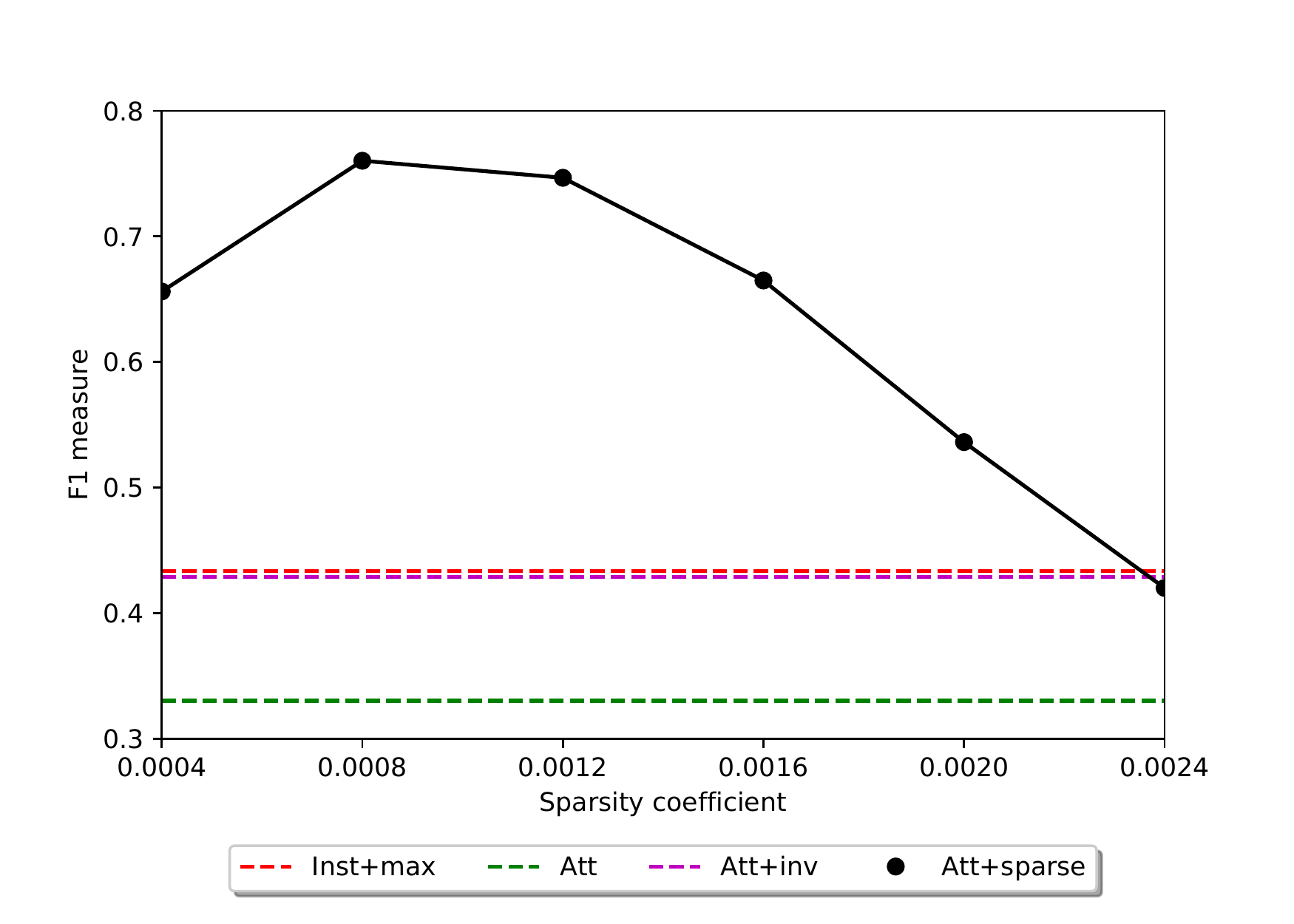}%
% \caption{Simulation results for the network.}
\label{colon:f1}}

\hfil
\subfloat[F1 measure for BREAST CANCER.]{\includegraphics[width=1.0\linewidth]{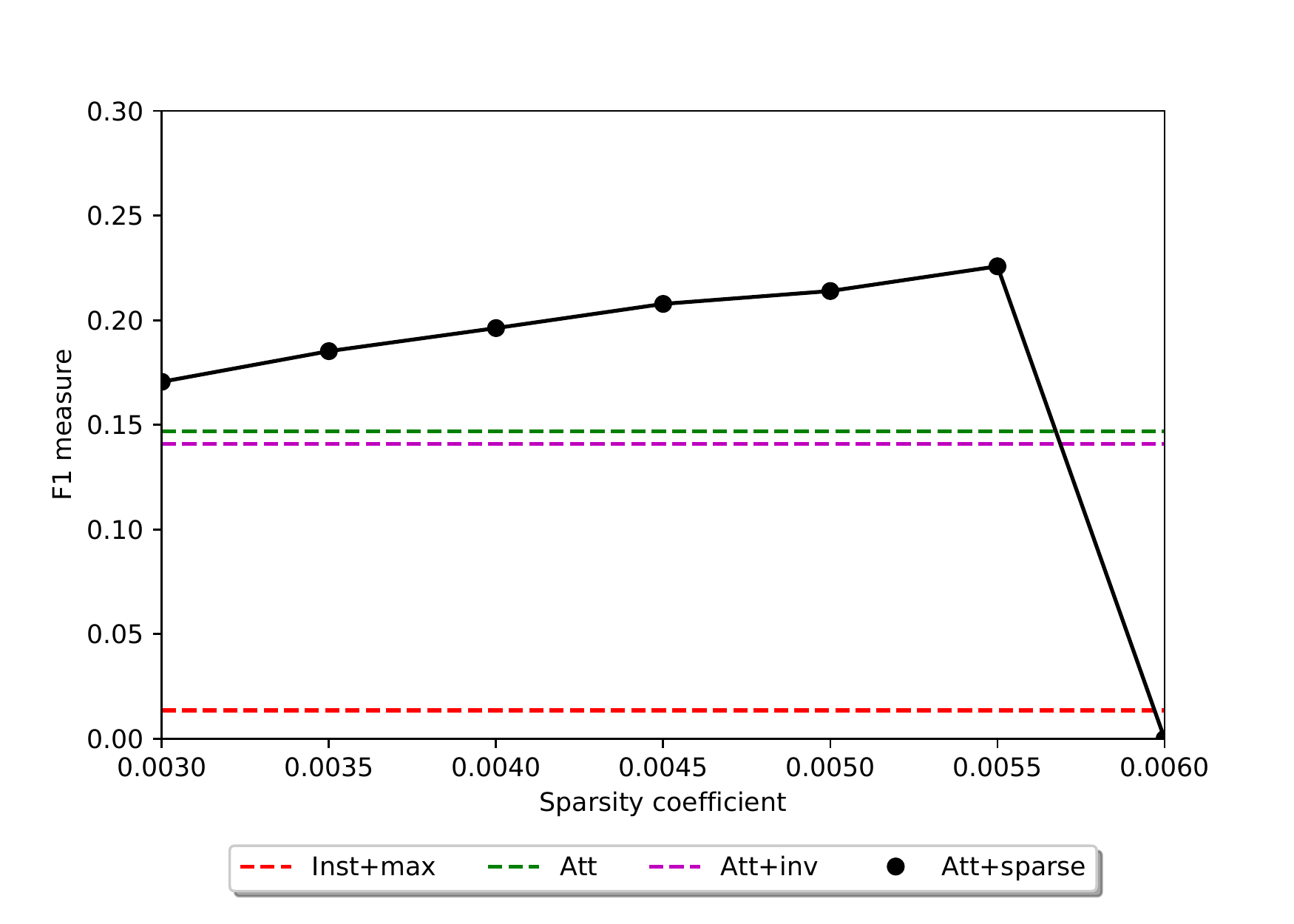}%
\label{breast:f1}}
\caption{F1 measure for MIL datasets.}
\end{figure}

\subsection{Results and analyses}

\begin{figure}[!t]
  \subfloat[All patches of an image.]{
   \begin{minipage}[c][0.9\width]{0.226\textwidth}
      \centering
      \includegraphics[width=1.\linewidth]{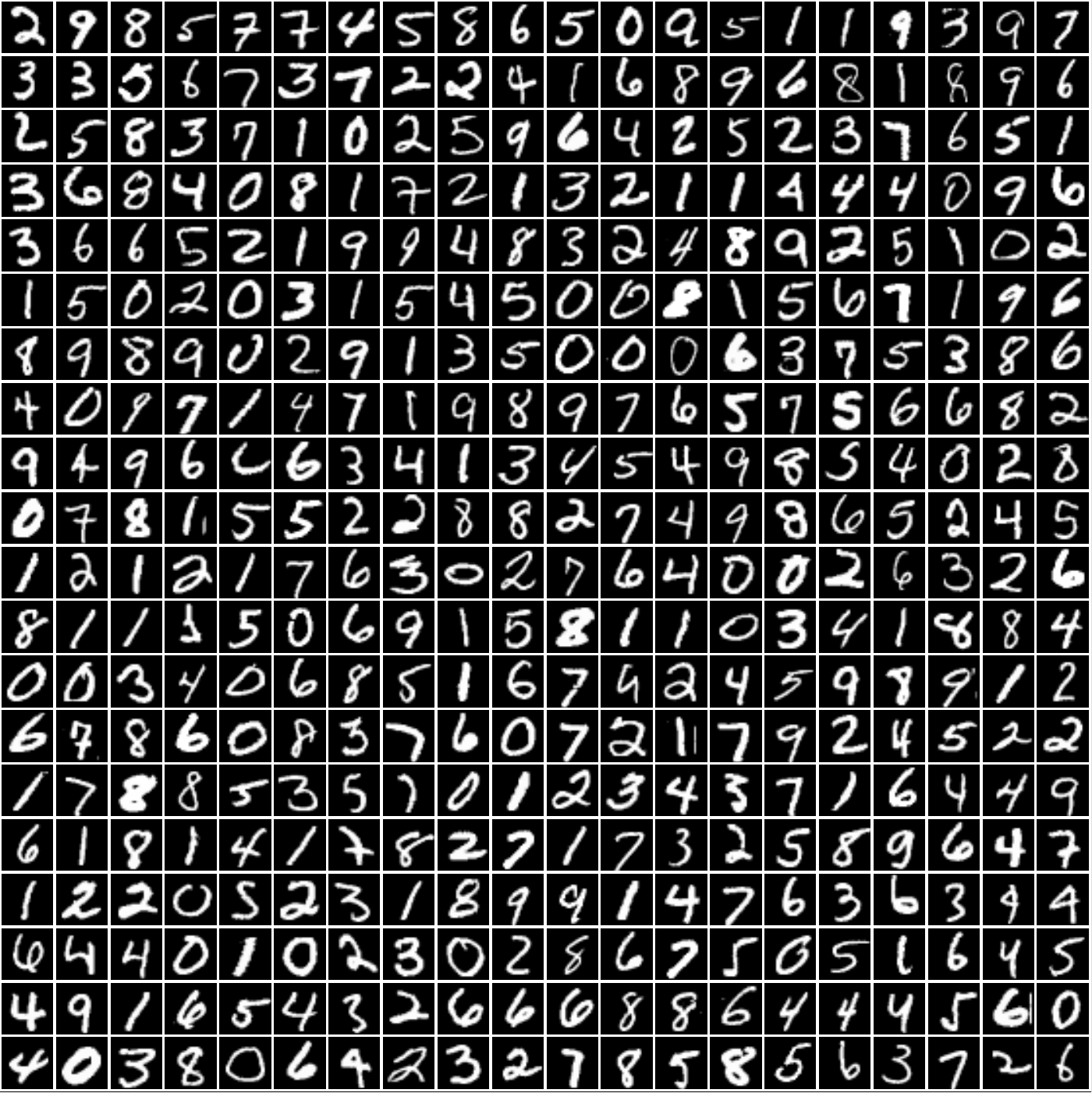}
   \end{minipage}}
  \subfloat[All refined patches.]{
   \begin{minipage}[c][0.9\width]{0.226\textwidth}
      \centering
      \includegraphics[width=1.\linewidth]{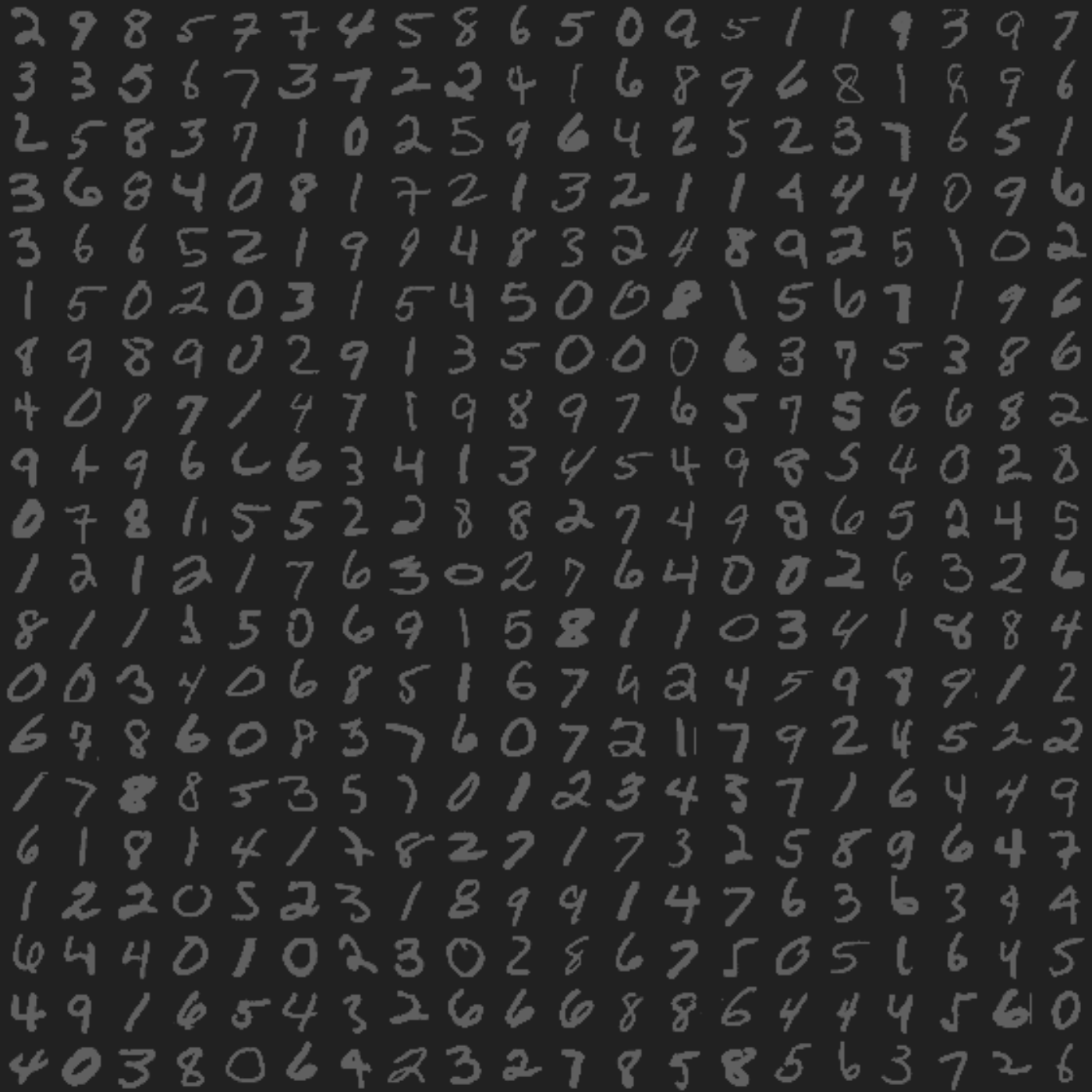}
   \end{minipage}}
 \hfill
  \subfloat[Heatmap of an attention-based deep MIL model.]{
   \begin{minipage}[c][0.9\width]{0.226\textwidth}
      \centering
      \includegraphics[width=1\linewidth]{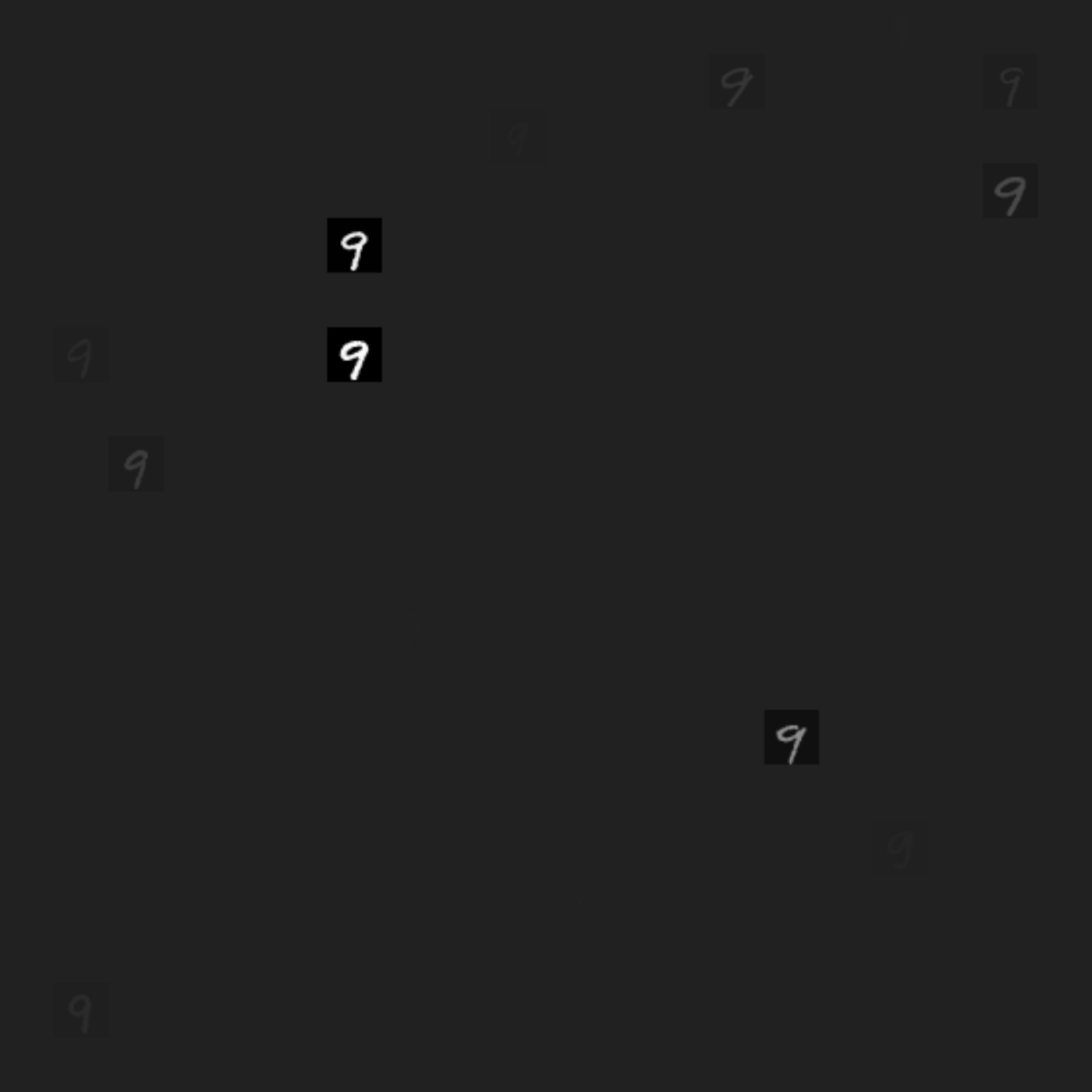}
   \end{minipage}}
  \subfloat[Heatmap of a sparse network inversion.]{
   \begin{minipage}[c][0.9\width]{0.226\textwidth}
      \centering
      \includegraphics[width=1.\linewidth]{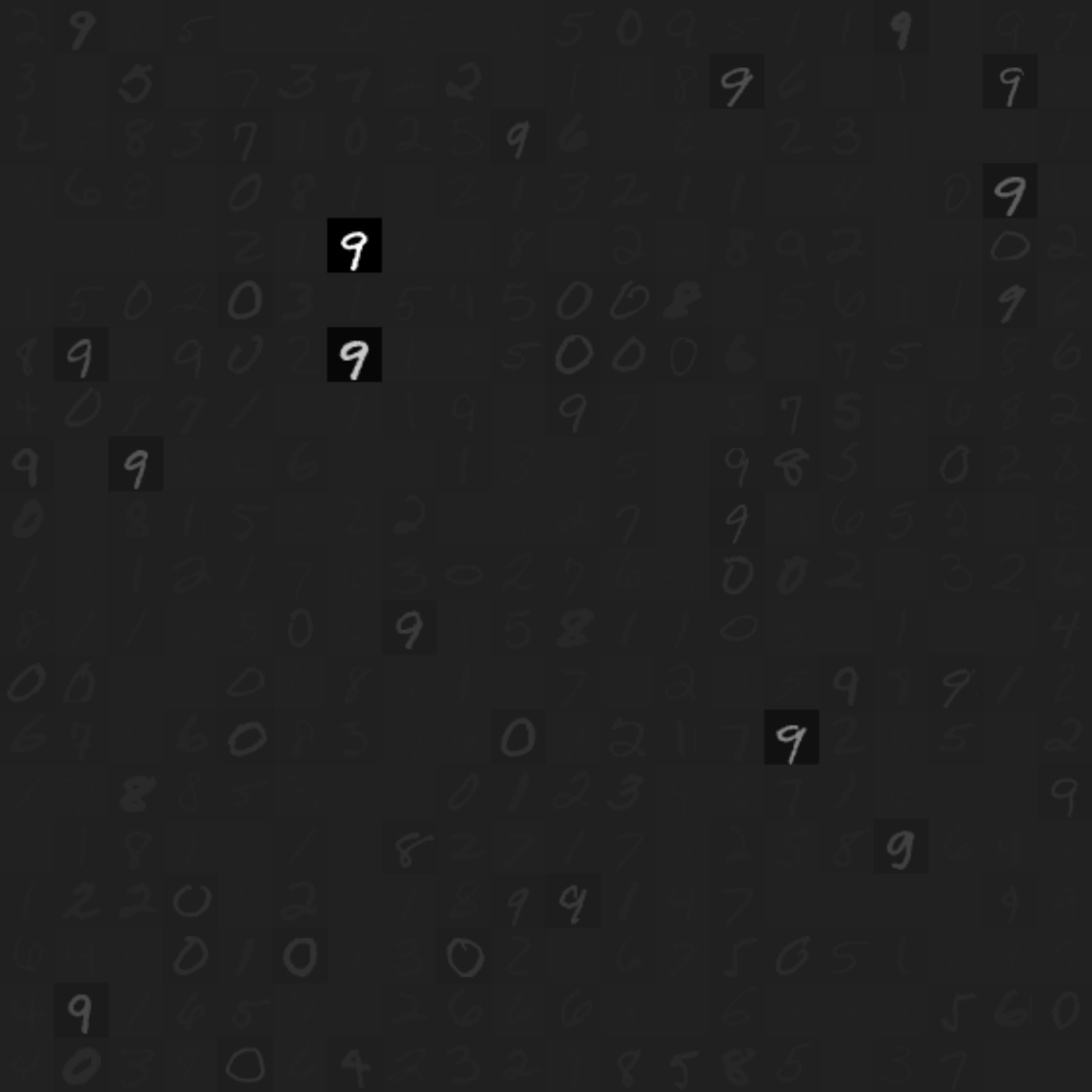}
   \end{minipage}}
 \hfill
  \subfloat[Visualization result of an attention-based deep MIL model.]{
   \begin{minipage}[c][0.9\width]{0.226\textwidth}
      \centering
      \includegraphics[width=1\linewidth]{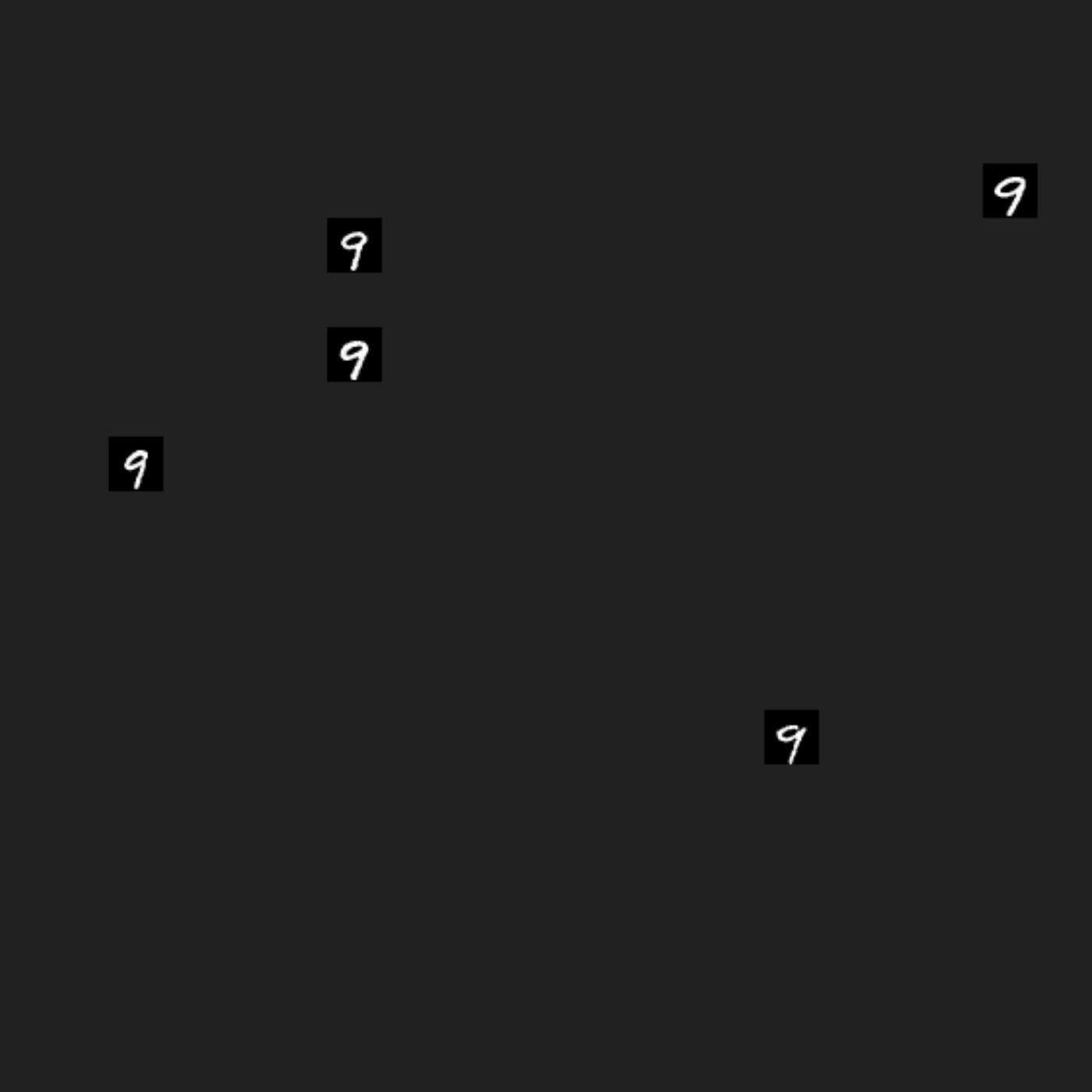}
   \end{minipage}}
  \subfloat[Visualization result of sparse network inversion.]{
   \begin{minipage}[c][0.9\width]{0.226\textwidth}
      \centering
      \includegraphics[width=1.\linewidth]{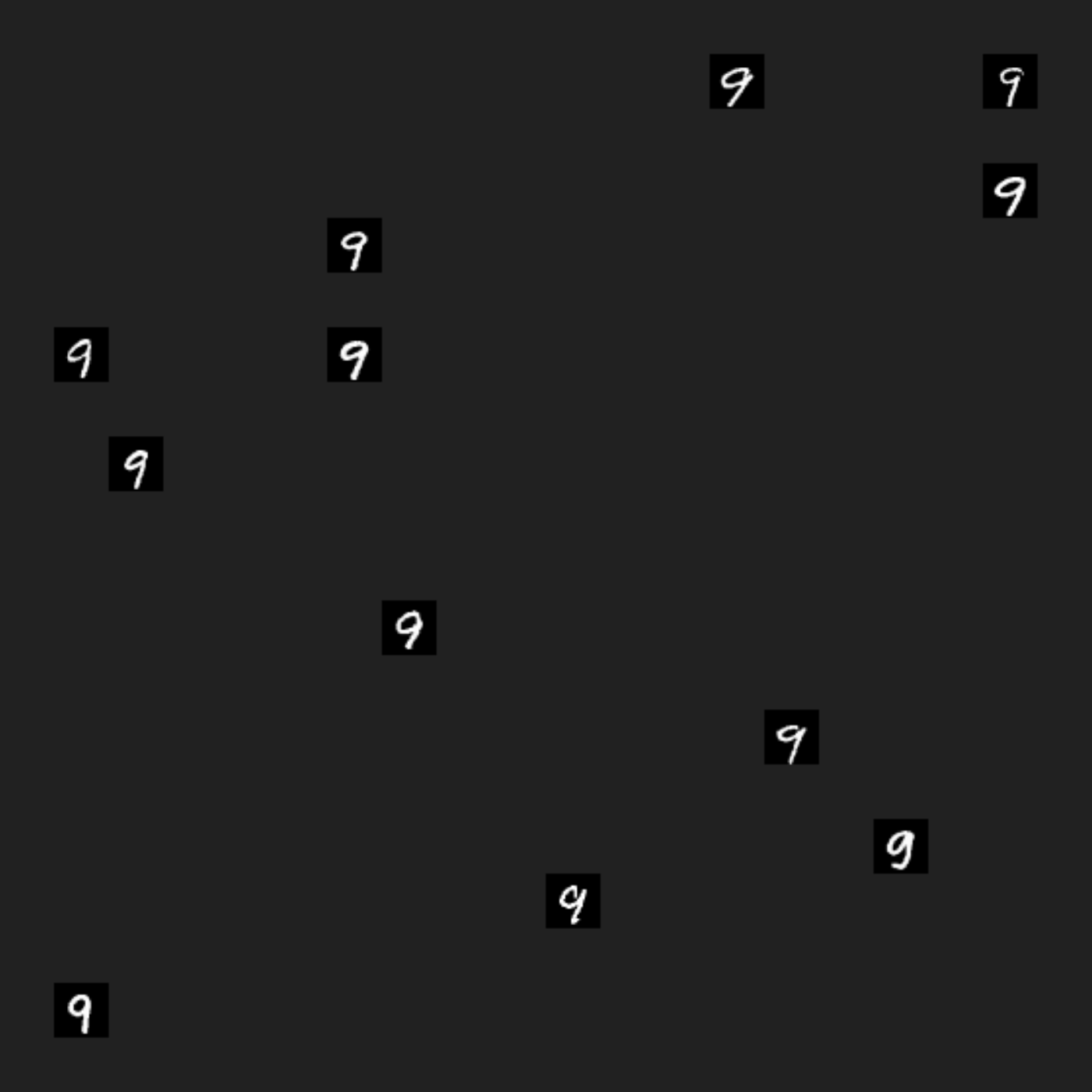}
   \end{minipage}}
 \hfill
 \centering
  \subfloat[Ground truth.]{
   \begin{minipage}[c][0.9\width]{0.226\textwidth}
      \centering
      \includegraphics[width=1.\linewidth]{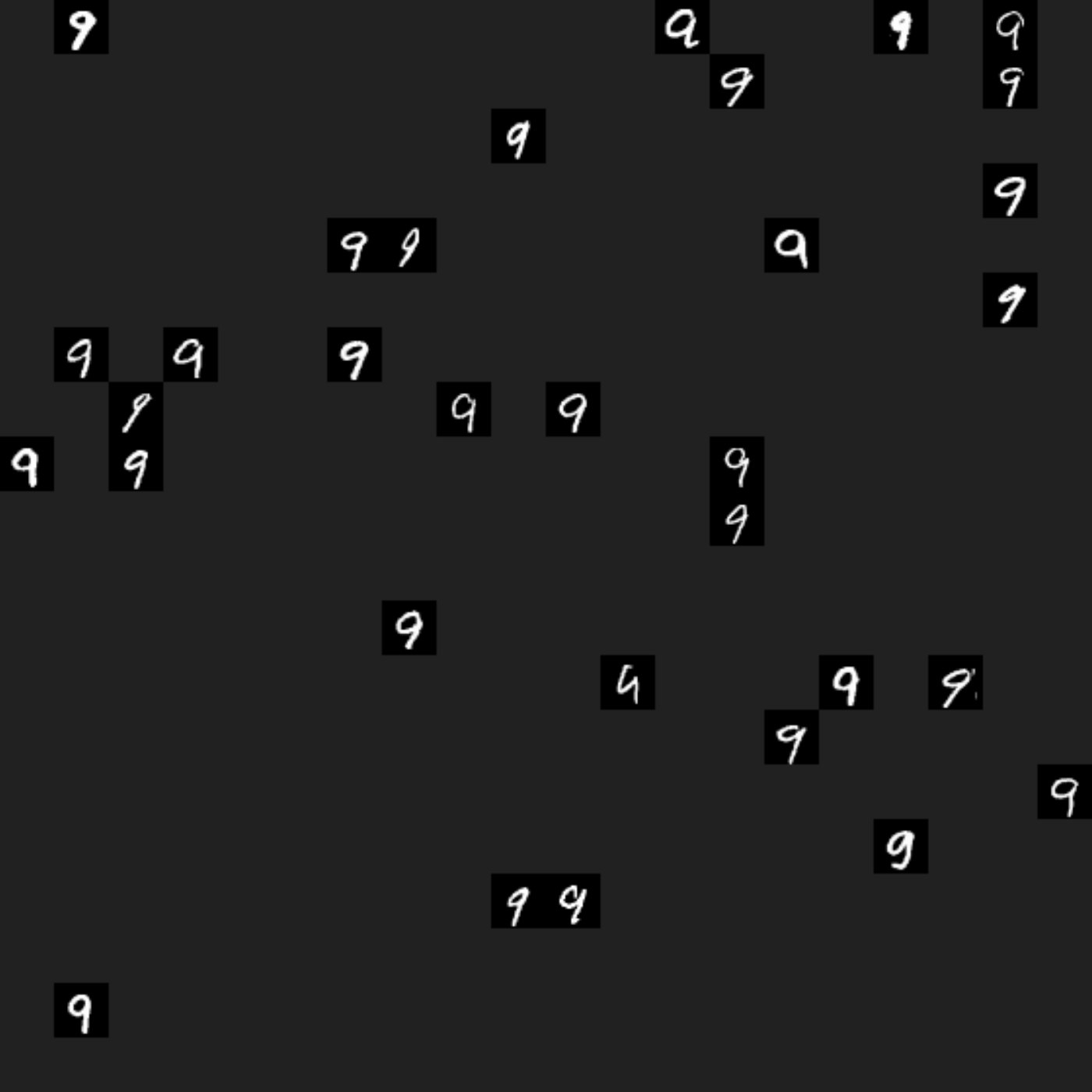}
   \end{minipage}}
\caption{Visualization results of an image in MNIST-based image MIL dataset.}

\end{figure}

\begin{figure}[!t]
  \subfloat[All patches of an image.]{
   \begin{minipage}[c][0.9\width]{0.226\textwidth}
      \centering
      \includegraphics[width=1.\linewidth]{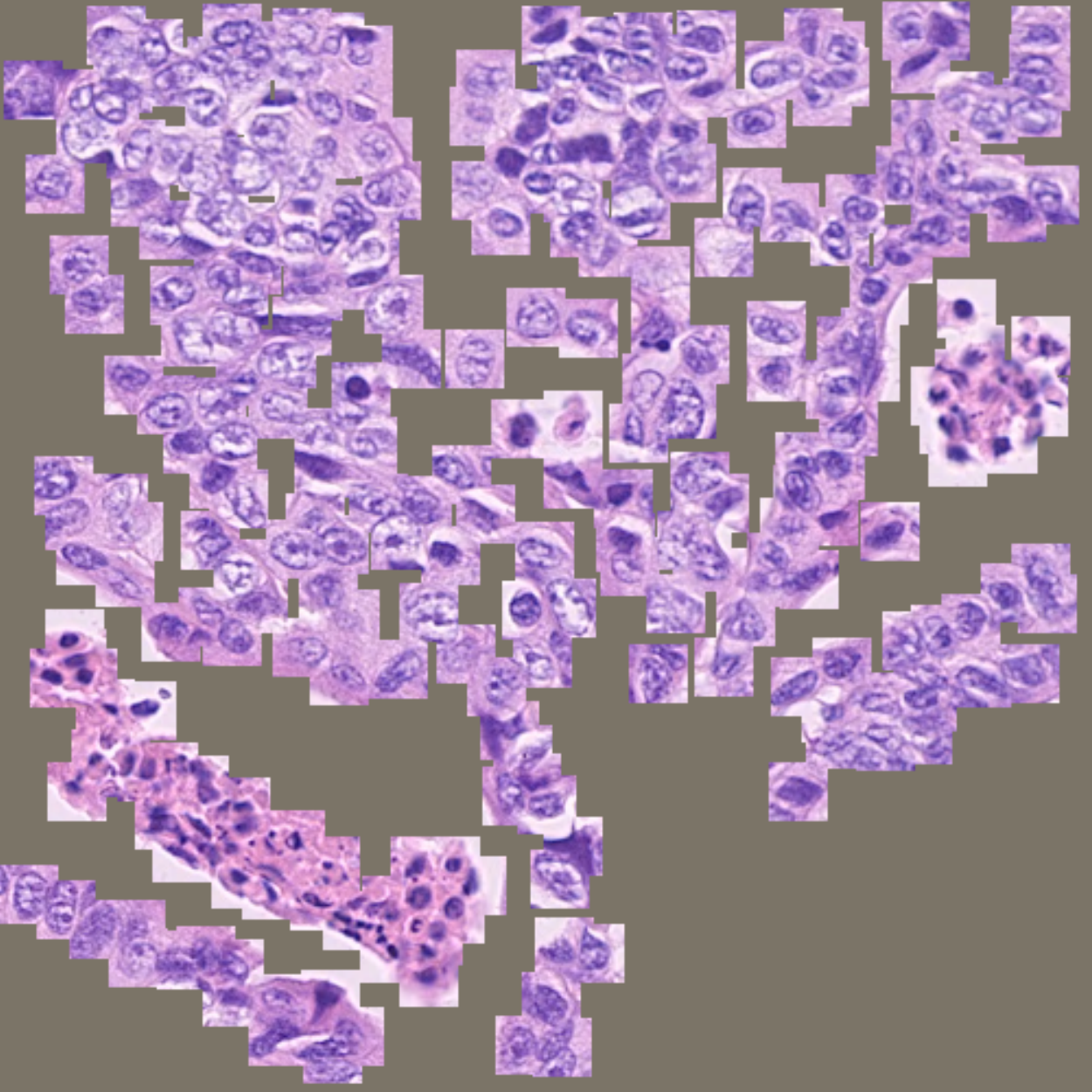}
   \end{minipage}}
  \subfloat[All refined patches.]{
   \begin{minipage}[c][0.9\width]{0.226\textwidth}
      \centering
      \includegraphics[width=1.\linewidth]{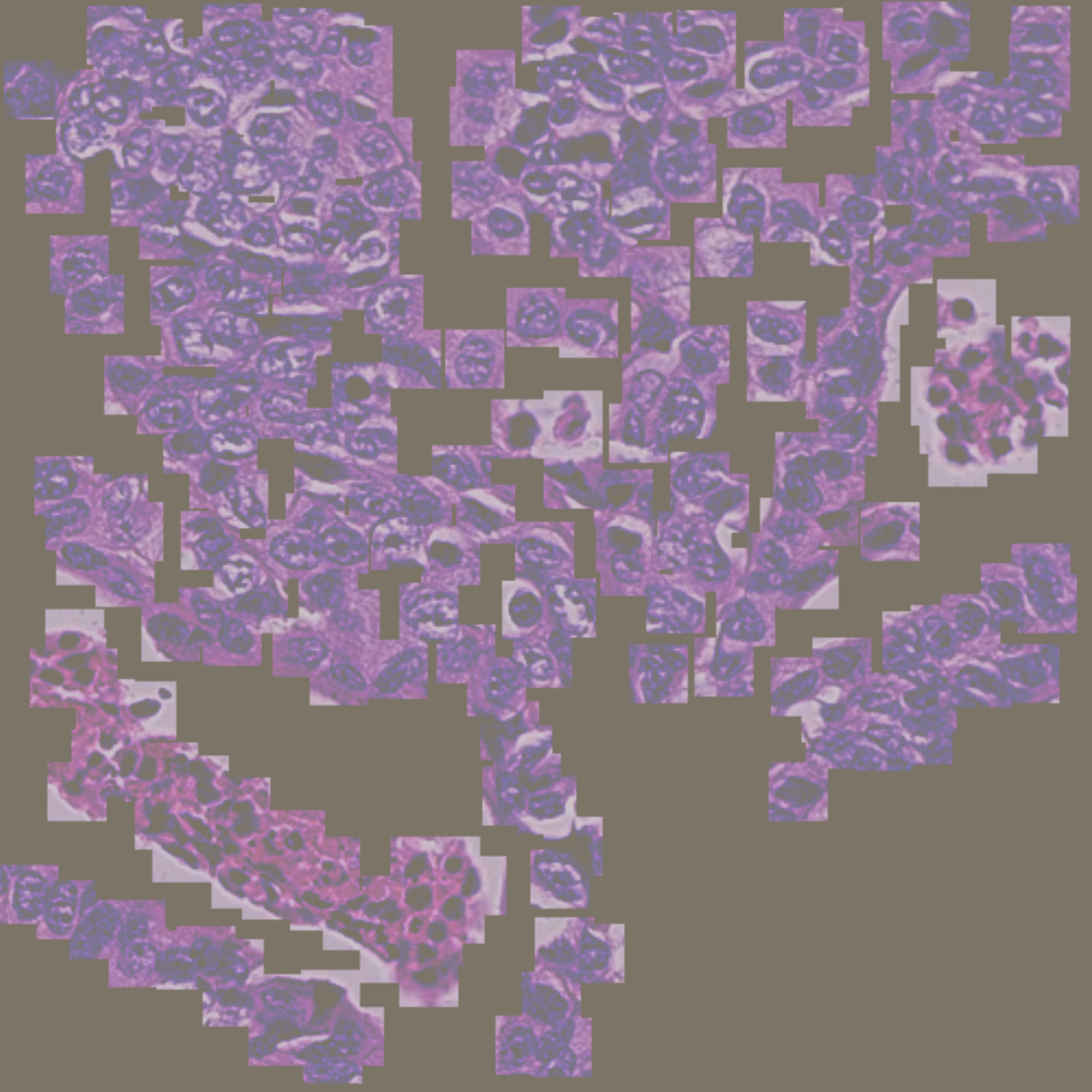}
   \end{minipage}}
 \hfill
  \subfloat[Heatmap of an attention-based deep MIL model.]{
   \begin{minipage}[c][0.9\width]{0.226\textwidth}
      \centering
      \includegraphics[width=1\linewidth]{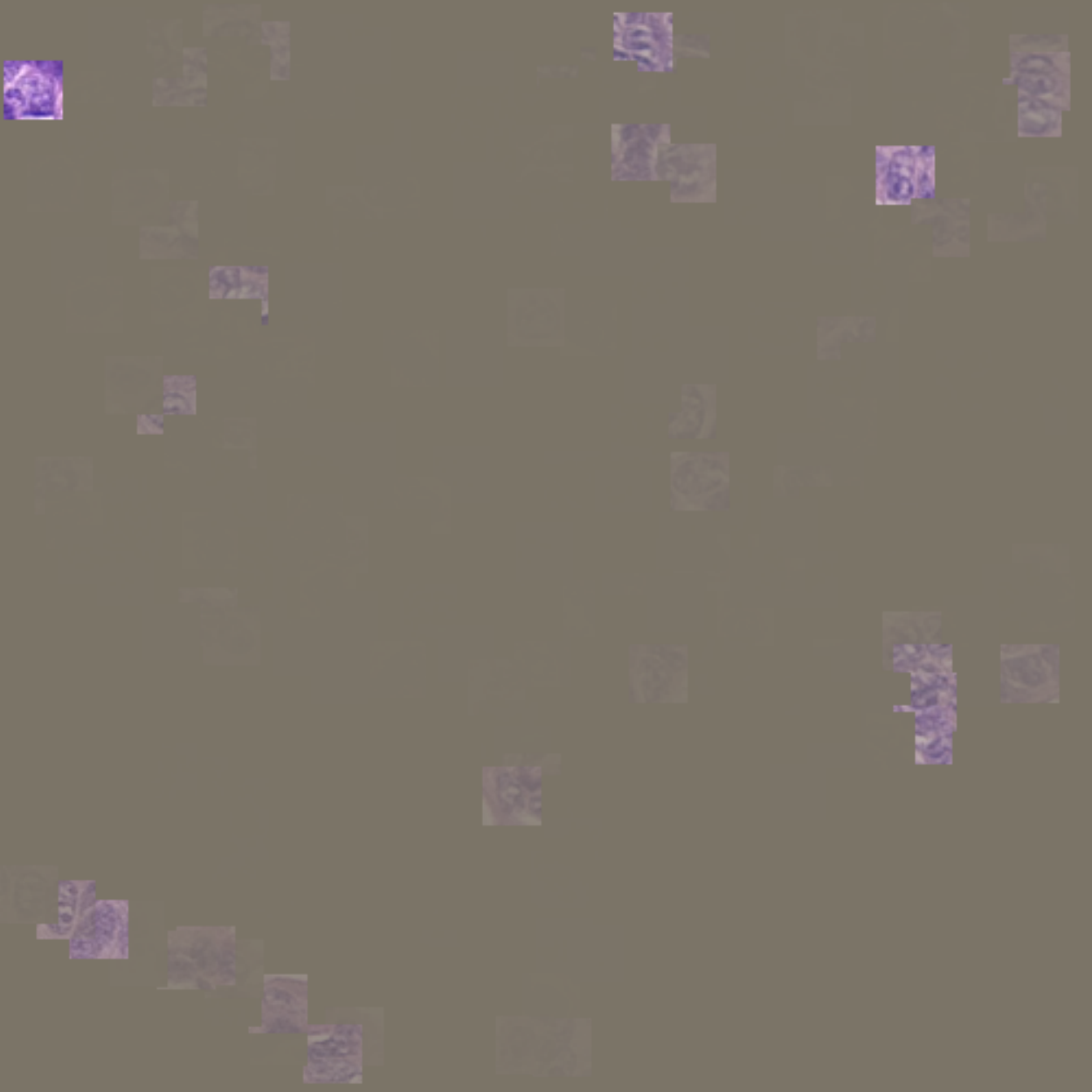}
   \end{minipage}}
  \subfloat[Heatmap of a sparse network inversion.]{
   \begin{minipage}[c][0.9\width]{0.226\textwidth}
      \centering
      \includegraphics[width=1.\linewidth]{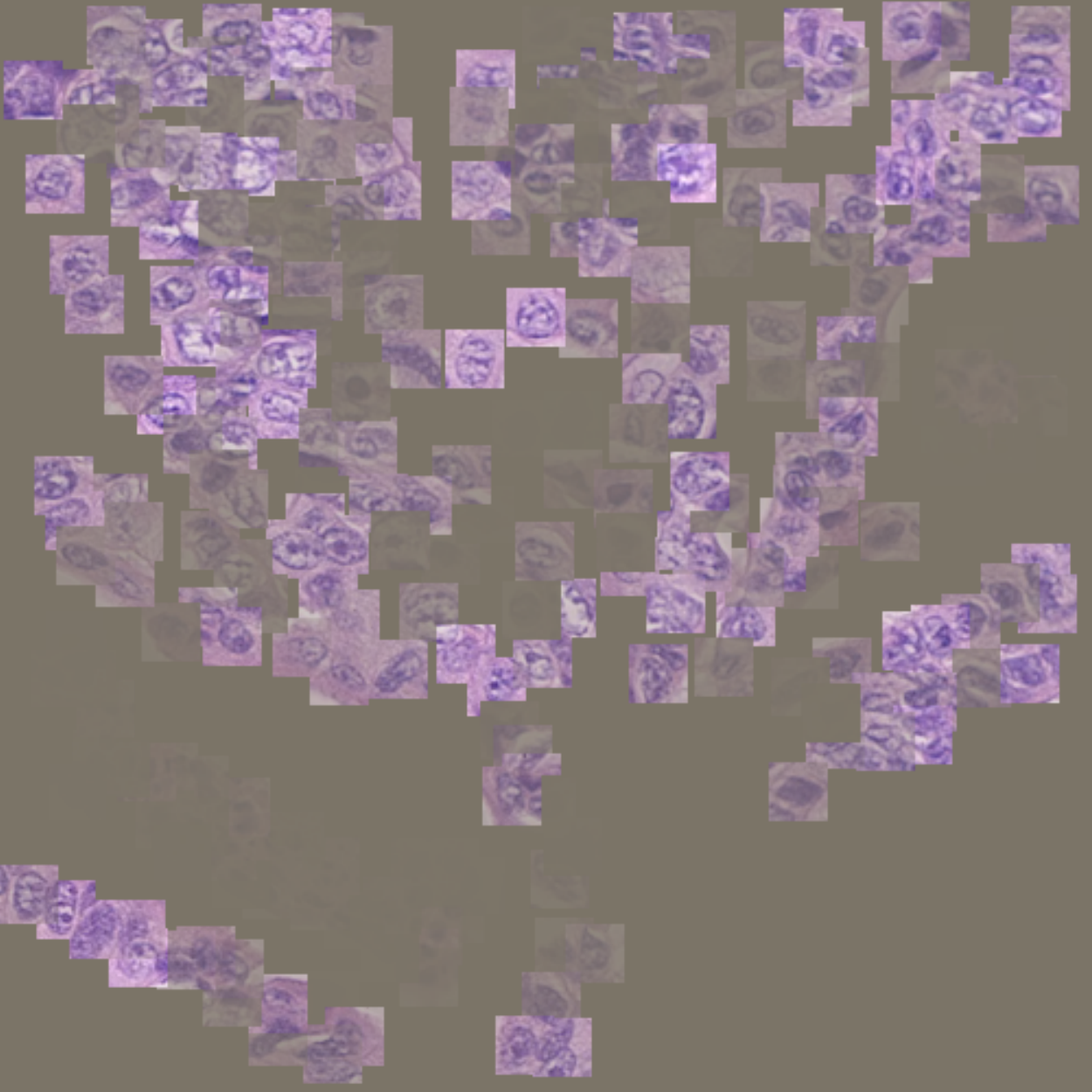}
   \end{minipage}}
 \hfill
  \subfloat[Visualization result of an attention-based deep MIL model.]{
   \begin{minipage}[c][0.9\width]{0.226\textwidth}
      \centering
      \includegraphics[width=1\linewidth]{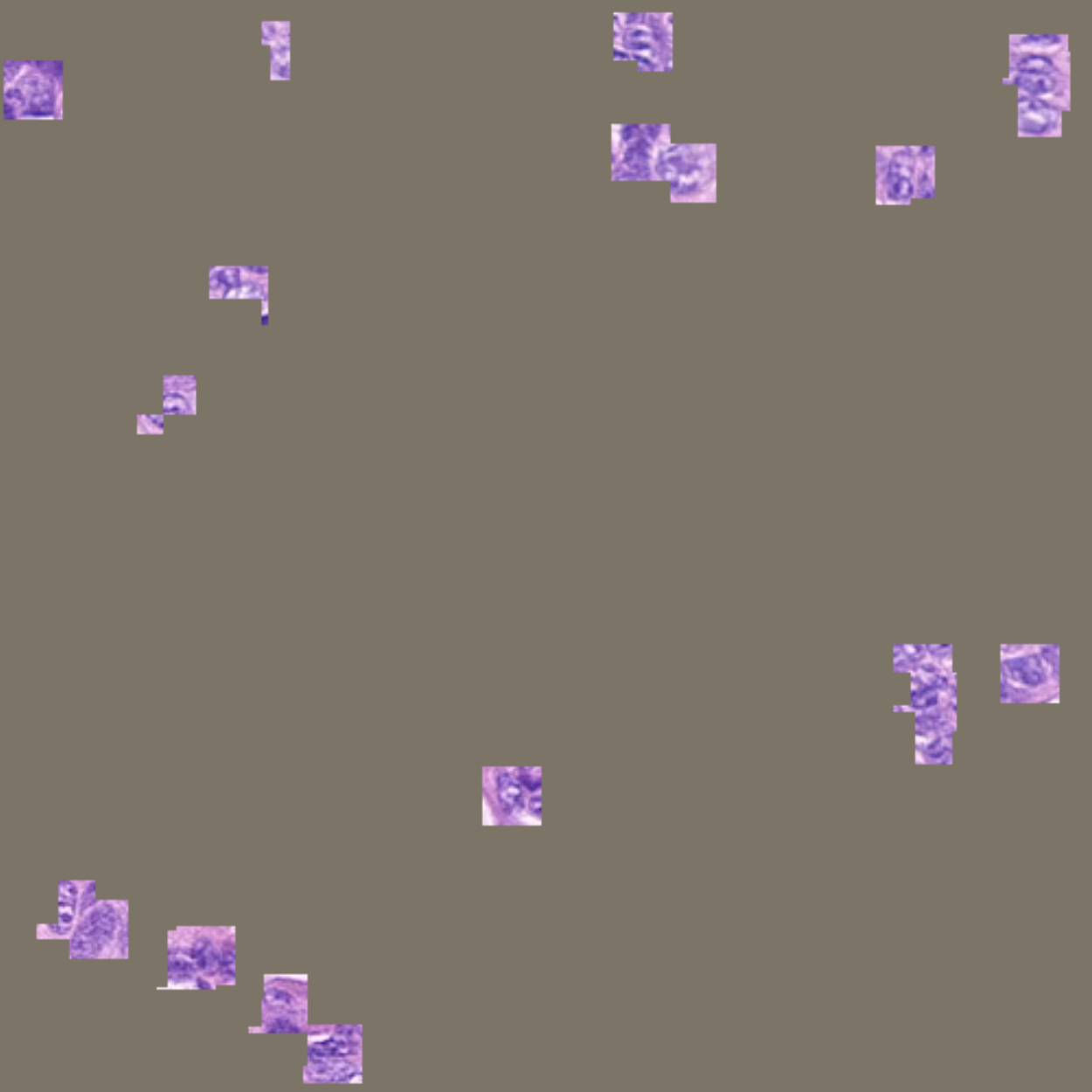}
   \end{minipage}}
  \subfloat[Visualization result of sparse network inversion.]{
   \begin{minipage}[c][0.9\width]{0.226\textwidth}
      \centering
      \includegraphics[width=1.\linewidth]{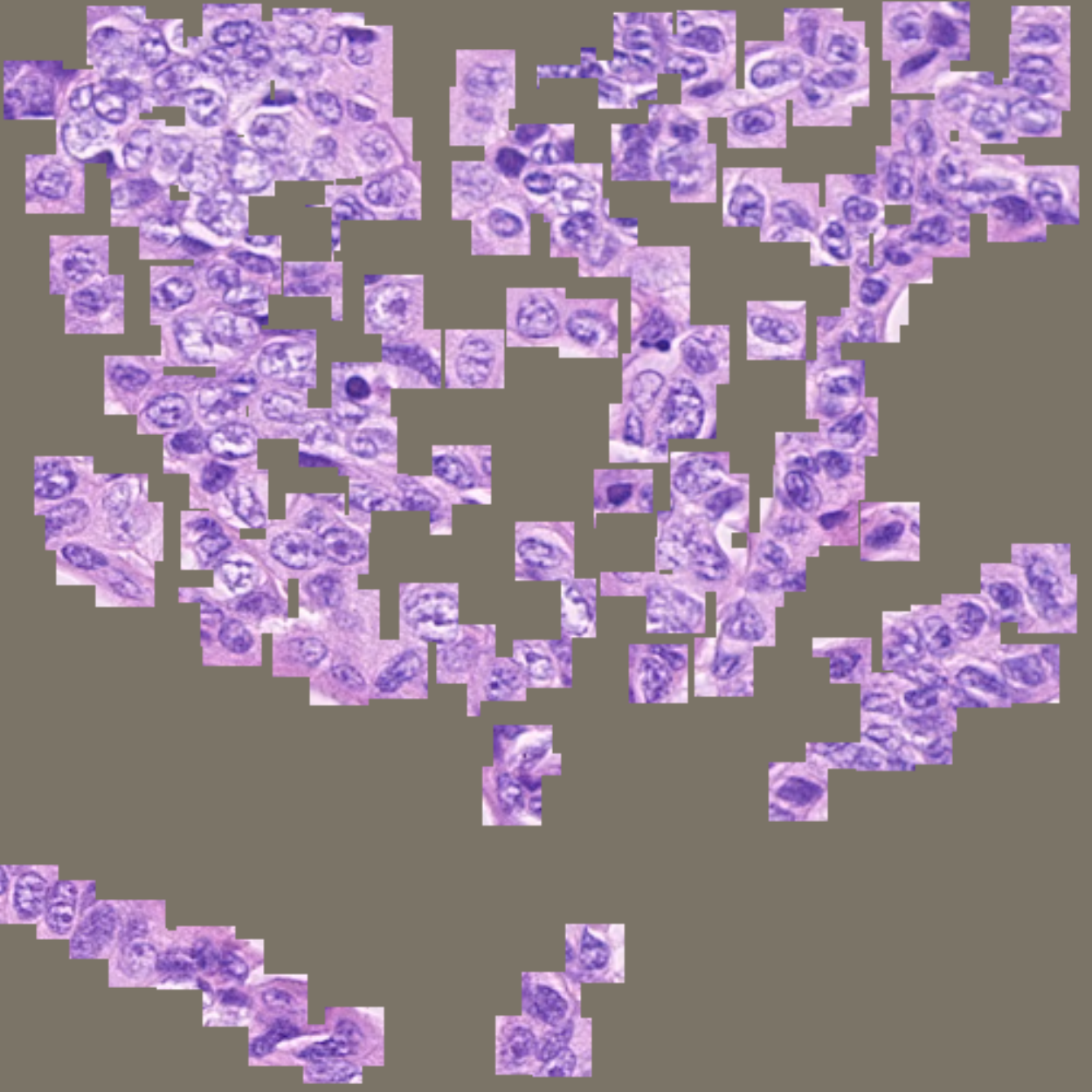}
   \end{minipage}}
 \hfill
 \centering
  \subfloat[Ground truth.]{
   \begin{minipage}[c][0.9\width]{0.226\textwidth}
      \centering
      \includegraphics[width=1.\linewidth]{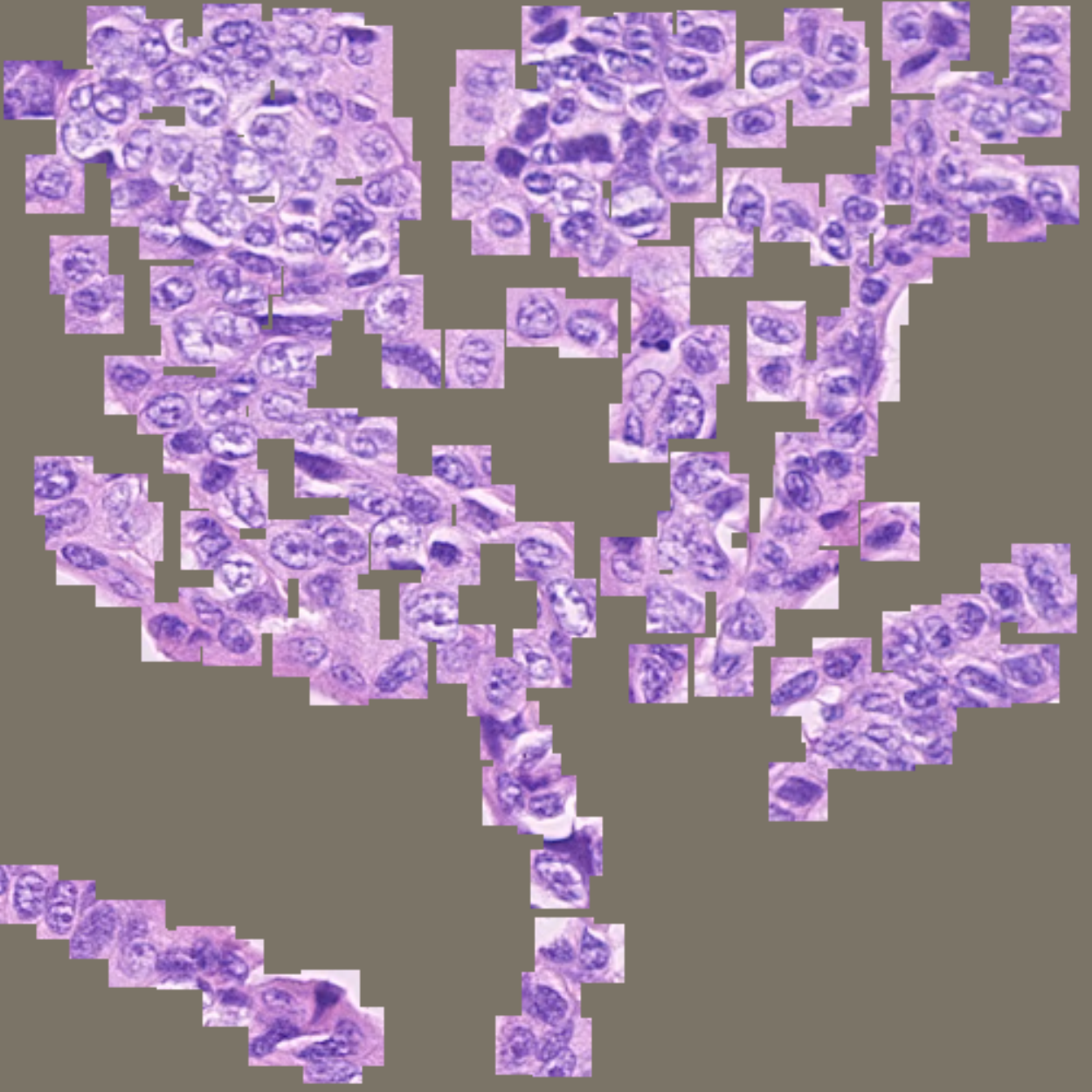}
   \end{minipage}}
\caption{Visualization results of an image in COLON CANCER.}

\end{figure}

% \begin{figure}[!t]
%   \subfloat[All patches of an image.]{
%   \begin{minipage}[c][0.9\width]{0.226\textwidth}
%       \centering
%       \includegraphics[width=1.\linewidth]{51_all.pdf}
%   \end{minipage}}
%   \subfloat[All refined patches.]{
%   \begin{minipage}[c][0.9\width]{0.226\textwidth}
%       \centering
%       \includegraphics[width=1.\linewidth]{51_inversion.pdf}
%   \end{minipage}}
%  \hfill
%   \subfloat[Heatmap of an attention-based deep MIL model.]{
%   \begin{minipage}[c][0.9\width]{0.226\textwidth}
%       \centering
%       \includegraphics[width=1\linewidth]{51_att_heatmap.pdf}
%   \end{minipage}}
%   \subfloat[Heatmap of a sparse network inversion.]{
%   \begin{minipage}[c][0.9\width]{0.226\textwidth}
%       \centering
%       \includegraphics[width=1.\linewidth]{51_inv_heatmap.pdf}
%   \end{minipage}}
%  \hfill
%   \subfloat[Visualization result of an attention-based deep MIL model.]{
%   \begin{minipage}[c][0.9\width]{0.226\textwidth}
%       \centering
%       \includegraphics[width=1\linewidth]{51_att.pdf}
%   \end{minipage}}
%   \subfloat[Visualization result of sparse network inversion.]{
%   \begin{minipage}[c][0.9\width]{0.226\textwidth}
%       \centering
%       \includegraphics[width=1.\linewidth]{51_inv.pdf}
%   \end{minipage}}
%  \hfill
%   \centering
%   \subfloat[Ground truth.]{
%   \begin{minipage}[c][0.9\width]{0.226\textwidth}
%       \centering
%       \includegraphics[width=1.\linewidth]{51_gt.pdf}
%   \end{minipage}}
% \caption{Visualization results of an image in COLON CANCER.}
% \end{figure}

The bag-level classification results and KID results on all datasets are in Table 3 and Fig. 2, respectively. For bag-level classification performance, the attention-based deep MIL model outperforms the other MIL models which are based on instance-space paradigm for all datasets. On the other hand, in the KID performance, the attention-based deep MIL model shows a worse performance than the other model based on instance-space paradigm for MNIST-based image MIL dataset and COLON CANCER. In the case of BREAST CANCER, although the attention-based deep MIL model has better performance than the other model based on instance-space paradigm, the absolute performance is not optimal. Thus, despite the superior bag-level classification performance of the attention-based deep MIL model, there is a problem in using an attention-based deep MIL model for KID. However, if we apply neural network inversion or sparse network inversion to data in the attention-based deep MIL model, since these methods relieve the problem that attention-based deep MIL model focuses only on few distinguishable key instances by removing the constraint that data cannot be changed, the KID performance is improved. Especially, in the case of sparse network inversion, the KID performance is significantly improved as the effects of regularization and sparseness have a positive effect on the result.

To intuitively show that our method can provide better interpretable results than the attention-based deep MIL model, we provide a visualization of the results in Fig. 3 and Fig. 4, where Fig. 3 and Fig. 4 are related to MNIST-based image MIL dataset and COLON CANCER, respectively. Each figure consists of seven images: (a) all patches of an image in the each dataset; (b) all refined patches that are the result of sparse network inversion; (c) heatmap of the attention-based deep MIL model: every patch from (a) multiplied by its corresponding Min-Max normalized attention score of every patch from (a); (d) heatmap of the sparse network inversion: every patch from (a) multiplied by its corresponding Min-Max normalized attention score of every patch from (b); (e) visualization result of the attention-based deep MIL model: every patch from (a) multiplied by its corresponding value that applies threshold to Min-Max normalized attention score of every patch from (a); (f) visualization result of sparse network inversion: every patch from (a) multiplied by its corresponding value that applies threshold to Min-Max normalized attention score of every patch from (b); (g) ground truth of the patches in instance-level. As shown in the Fig. 3-(c) and Fig. 4-(c), Min-Max normalized attention scores of the attention-based deep MIL model are skewed to few distinguishable key instances. However, if we apply our method to the patches, Min-Max normalized attention scores are scattered around more key instances than before applying sparse network inversion to the patches as shown in the Fig. 3-(d) and Fig. 4-(d). By applying threshold to the normalized attention scores of the attention-based deep MIL model and sparse network inversion, we can perform KID as shown in the Fig. 3-(e), Fig. 3-(f), Fig. 4-(e), and Fig. 4-(f). From these results, although we use only bag-level data and bag-level label during training, we can confirm that the attention-based deep MIL model finds key instances in a positive bag. However, if we just use the attention-based deep MIL model as shown in Fig. 3-(e) and Fig. 4-(e), the model finds only few key instances. On the other hand, as shown in Fig. 3-(f) and Fig. 4-(f), our method allows the model to find more key instances than before applying our method.

From these experiments on the MNIST-based image MIL dataset and two real-world histopathology datasets, we can see that our method significantly improves the KID performance while the bag-level classification performance is maintained. This means that applying sparse network inversion to the attention-based deep MIL model helps improve the KID performance of the attention-based deep MIL model.

\section{Conclusion}
In this paper, we propose a method to maintain the performance of bag-level classification and to improve KID performance of an attention-based deep MIL model. Our method applies sparse network inversion to the MIL model. In experiments, we measured the accuracy of bag-level classification and F1 measure of instance-level classification for our proposed method for an MNIST-based image MIL dataset and two histopathology datasets, and our proposed method significantly improved the KID performance of the attention-based deep MIL model. The bag-level decision of the MIL model can be interpreted with the key instances. Therefore, our research will be useful in area such as medicine where the interpretation of the model's behavior can be important.

% conference papers do not normally have an appendix

% use section* for acknowledgment
\section*{Acknowledgment}
This work was supported by Institute for Information \& communications Technology Promotion(IITP) grant funded by the Korea government(MSIT) (IITP-2018-0-00584), and Samsung Electronics Co., Ltd.

%The authors would like to thank...

% trigger a \newpage just before the given reference
% number - used to balance the columns on the last page
% adjust value as needed - may need to be readjusted if
% the document is modified later
%\IEEEtriggeratref{8}
% The "triggered" command can be changed if desired:
%\IEEEtriggercmd{\enlargethispage{-5in}}

% references section

% can use a bibliography generated by BibTeX as a .bbl file
% BibTeX documentation can be easily obtained at:
% http://mirror.ctan.org/biblio/bibtex/contrib/doc/
% The IEEEtran BibTeX style support page is at:
% http://www.michaelshell.org/tex/ieeetran/bibtex/
%\bibliographystyle{IEEEtran}
% argument is your BibTeX string definitions and bibliography database(s)
%\bibliography{IEEEabrv,../bib/paper}
%
% <OR> manually copy in the resultant .bbl file
% set second argument of \begin to the number of references
% (used to reserve space for the reference number labels box)

{\small
\bibliographystyle{IEEEtran}
%\bibliography{IEEEabrv,ICPR}
\bibliography{ICPR}
}

% \begin{thebibliography}{1}

% \bibitem{IEEEhowto:kopka}

% H.~Kopka and P.~W. Daly, \emph{A Guide to \LaTeX}, 3rd~ed.\hskip 1em plus
%   0.5em minus 0.4em\relax Harlow, England: Addison-Wesley, 1999.

% \end{thebibliography}

% that's all folks
\end{document}